\documentclass{article}

\usepackage[preprint, nonatbib]{neurips_2023}

\usepackage[utf8]{inputenc} 
\usepackage[T1]{fontenc}    
\usepackage{hyperref}       
\usepackage{url}            
\usepackage{booktabs}       
\usepackage{amsfonts}       
\usepackage{nicefrac}       
\usepackage{microtype}      
\usepackage{xcolor}         

\usepackage[
    natbib=true,
    citestyle=numeric-comp,
    bibstyle=ieee,
    sorting=nty
]{biblatex}
\usepackage{graphicx}
\usepackage{caption}
\usepackage{subcaption}
\usepackage{bm}
\usepackage{amsmath,amssymb}
\usepackage{booktabs}
\usepackage{algorithm}
\usepackage[noend]{algorithmic}
\usepackage{multirow}
\usepackage[flushleft]{threeparttable}

\usepackage{makecell}
\usepackage{multirow}
\usepackage{xspace}
\usepackage{color}
\usepackage{wrapfig}
\usepackage{bbding}

\newcommand{\ie}{\emph{i.e.,}\xspace}
\newcommand{\eg}{\emph{e.g.,}\xspace}

\usepackage{siunitx}
\usepackage{dcolumn}

\usepackage{calc}
\newcommand{\tokens}[1]{\colorbox{blue!30}{\parbox[b][0.6em]{\widthof{#1}}{#1}}}

\definecolor{YELLOW}{rgb}{0.83, 0.61, 0.18}

\title{OpenMoE: An Early Effort on Open Mixture-of-Experts Language Models}

\stepcounter{footnote}

\author{
Fuzhao Xue$^{1}$\thanks{Email: f.xue@u.nus.edu} \And
Zian Zheng$^{1}$ \And
Yao Fu$^2$ \And
Jinjie Ni$^{1}$ \And
Zangwei Zheng$^{1}$ \And
Wangchunshu Zhou$^3$ \And
Yang You$^{1}$ \AND \\
$^1$National University of Singapore \\
$^2$University of Edinburgh \\
$^3$ETH Zurich \\
}

\begin{document}

\maketitle

\begin{figure}[h]
  \begin{center}
    \includegraphics[width=0.3\textwidth]{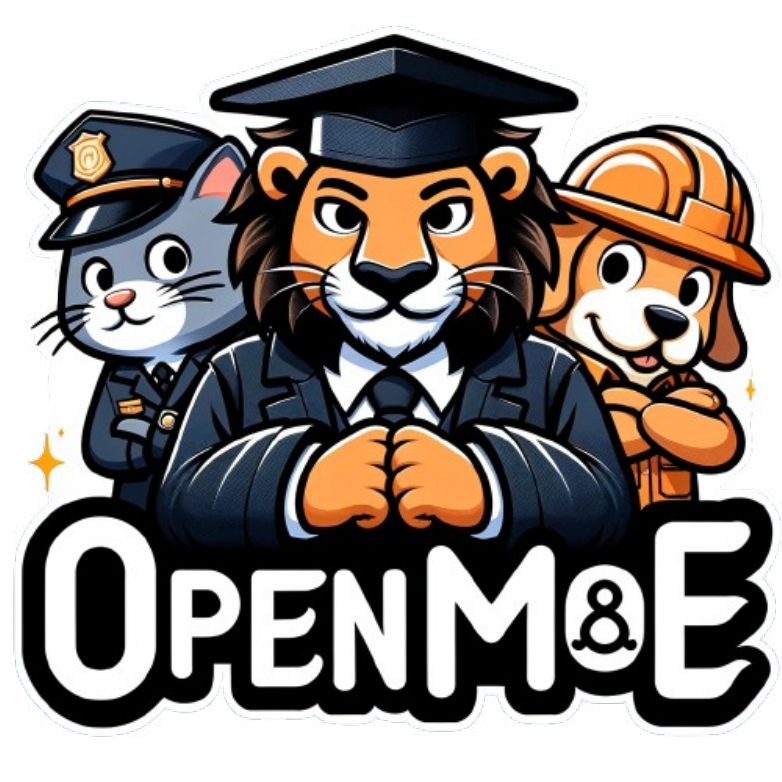}
  \end{center}
\end{figure}

\begin{abstract}

To help the open-source community have a better understanding of Mixture-of-Experts (MoE) based large language models (LLMs), we train and release OpenMoE, a series of fully open-sourced and reproducible decoder-only MoE LLMs, ranging from 650M to 34B parameters and trained on up to over 1T tokens. Our investigation confirms that MoE-based LLMs can offer a more favorable cost-effectiveness trade-off than dense LLMs, highlighting the potential effectiveness for future LLM development.

One more important contribution of this study is an in-depth analysis of the routing mechanisms within our OpenMoE models, leading to three significant findings: Context-Independent Specialization, Early Routing Learning, and Drop-towards-the-End.
We discovered that routing decisions in MoE models are predominantly based on token IDs, with minimal context relevance.
The token-to-expert assignments are determined early in the pre-training phase and remain largely unchanged. This imperfect routing can result in performance degradation, particularly in sequential tasks like multi-turn conversations, where tokens appearing later in a sequence are more likely to be dropped.
Finally, we rethink our design based on the above-mentioned observations and analysis. To facilitate future MoE LLM development, we propose potential strategies for mitigating the issues we found and further improving off-the-shelf MoE LLM designs.\footnote{https://github.com/XueFuzhao/OpenMoE}


\end{abstract}


\section{Introduction}\label{Introduction}


Large Language Model (LLM) has exhibited remarkable performance on various NLP tasks~\citep{2020t5,li2022self}, and has even become a part of our daily lives through chatbot applications such as ChatGPT, Bard, and Copilot. 
However, LLMs are computationally expensive in both training and inference.
As LLMs become increasingly prevalent, enhancing their performance without proportionally increasing computational resources is a critical challenge. 
In response to this challenge, \citet{riquelme2021scaling,fedus2021switch} proposed the Mixture-of-Experts (MoE) to scale up the trainable parameters of the transformer with little additional computation overhead. 
Recent advancements in MoE-based language models, such as GLaM~\cite{du2022glam} and ST-MoE~\cite{zoph2022st} have demonstrated superior performance in various tasks.
However, before the release of OpenMoE, there were few open-sourced MoE language models trained with trillion-level diverse datasets.


In this work, we set forth three primary goals: (1) To offer a first-attempt solution in detail for training a decoder-only MoE model within the existing framework of training LLMs. (2) To perform an in-depth analysis of the MoE routing mechanisms, thereby providing the research community with deeper insights into the behaviors and potential limitations of MoE-based LLMs. (3) To pave the way for future MoE LLM development. Through this early endeavor, we aim to stimulate and accelerate the growth of the open-source MoE community.

\textbf{Releasing OpenMoE.} First, we release OpenMoE, a series of open-sourced MoE-based LLMs, including:
(1) \underline{OpenMoE-Base/16E}: a small model with 0.65B parameters for debugging purposes. 16E means 16 experts per MoE layer;
(2) \underline{OpenMoE-8B/32E}: this variant features 8B parameters in total, activating around 2B parameters per token in Transformer blocks, and is pre-trained on over 1 trillion tokens; 
(3) \underline{OpenMoE-8B/32E-Chat}, a chat version of OpenMoE-8B/32E, fine-tuned with a 100K subset of the WildChat~\cite{anonymous2024inthewildchat} dataset;
(4) \underline{OpenMoE-34B/32E}: a larger scale model, activating 6B parameters per token in Transformer blocks and trained with 200B tokens, serving as a testament to the scalability of our approach.
Detailed configuration can be found in Appendix~\ref{sec:hyper-param}
Our OpenMoE-8B/32E models achieved comparable performance with OpenLLaMA-3B~\cite{openlm2023openllama} and TinyLLaMA-1.1B~\cite{zhang2024tinyllama}, two dense open LLMs used higher training cost. Notably, On the MT-Bench~\cite{zheng2023judging}, OpenMoE-8B/32E-Chat outperformed the two dense LLMs significantly on the single-turn conversation. 
In addition, we release 5 intermediate checkpoints of OpenMoE-8B/32E, each trained with 200B more tokens than the previous one, to support and encourage future research. Section~\ref{sec:design}~and~\ref{sec:training} will discuss the design, training details, and evaluation results of OpenMoE.

\textbf{Exploring Advanced Training Strategies.} As part of our research endeavor, we are committed to exploring more advanced Techniques in LLM training:
(1) Different from the common practice of training models on in-house or text-dominated open-sourced data, we train OpenMoE with a substantial proportion of code, constituting up to 52.25\% during the early stages of pre-training;
(2) Moving beyond the conventional next-token prediction training objective, we investigate UL2 training objective~\cite{tay2022unifying}, motivated by its proven effectiveness in previous work~\cite{anil2023palm} and its good alignment with coding data~\cite{bavarian2022efficient}.
We acknowledge that the performance of our model, while acceptable, does not significantly exceed our expectations, which may be attributed to some sub-optimal design choices. Nevertheless, we believe that this exploratory work offers substantial value to the open-source community, particularly in assessing the potential and effectiveness of these under-explored techniques.

\textbf{Studying MoE Routing In-depth.} While MoE is effective, there remains a lack of study on why MoE performs well. From a high level, MoE introduces more trainable parameters than its dense counterpart. To keep the FLOPs fixed when scaling the number of parameters, MoE applies a routing layer that sparsely and adaptively assigns each token to a few experts. This process of sparse expert selection is crucial to MoE's functionality. Unfortunately, despite existing pieces of literature briefly visualizing the routing decison~\cite{shazeer2017outrageously,lewis2021base,zoph2022st, mustafa2022multimodal, riquelme2021scaling}, we still don't have a clear understanding of how the router works and how the routing decision impacts the results in MoE models, especially for the post-ChatGPT LLMs trained on a mixture of datasets from diverse domains.
In this work, we study this problem based on various taxonomies, including domain, language, task, and token. Our key findings are as follows: 
(1) \textbf{Context-independent Specialization}: MoE tends to simply cluster tokens based on similar token-level semantics, implying that, regardless of context, a certain token is more likely to be routed to a certain expert; (2) \textbf{Early Routing Learning}: Token ID routing specialization is established early in pre-training and remains largely fixed, resulting in tokens being consistently processed by the same experts throughout the training; (3) \textbf{Drop-towards-the-End}: Since each expert has a fixed max capacity, tokens appearing later in the sequence face a higher risk of being dropped if the expert is already at capacity. This issue is more severe in instruction-tuning datasets. These datasets often exhibit a domain gap compared to the pre-training data, meaning that the balanced token assignment strategies established and solidified during early pre-training may not be samely effective in instruction-tuning scenarios.
This is concerning as instruction data plays an important role in deploying LLMs to real-world applications. Section~\ref{sec:analysis} discusses the above phenomenons in detail.

\textbf{Rethinking Our Mistakes and Proposing Potential Solutions.} In retrospect, our project encountered several mistakes and made sub-optimal decisions (\eg aggressive data mixture), as detailed in Section~\ref{sec:rethink}. 
As an early open-source effort, we believe that sharing these experiences and insights is crucial, perhaps even more important than solely focusing on successful strategies. 
Based on our empirical findings during training and subsequent visualization analysis (Section~\ref{sec:analysis}), we have developed a set of potential solutions. We sincerely hope these insights can help the community develop better models in the future. 

The structure of this paper mirrors the lifecycle of the OpenMoE project, encompassing all its phases. This includes the initial design (Section~\ref{sec:design}), training and evaluation (Section~\ref{sec:training}, in-depth analysis (Section~\ref{sec:analysis}), and a rethinking of the OpenMoE project (Section~\ref{sec:rethink}).


\section{Designing OpenMoE}
\label{sec:design}

First, we introduce our initialized design of OpenMoE models regarding the pre-training data, model architecture, training objective, and supervised fine-tuning data. 

\subsection{Pre-training Dataset: More Code than Usual}

Modern LLMs are usually trained by a combination of datasets from various domains, \ie data mixture~\cite{brown2020language, rae2021scaling, hoffmann2022training, chowdhery2022palm, touvron2023llama}. Except for the LLMs customized towards coding (\eg StarCoder~\cite{li2023starcoder}, CodeLLaMA~\cite{roziere2023code}), most existing models' pre-training data is dominated by text data. For instance, the sampling rate of the GitHub dataset is only 4.5\% for LLaMA~\cite{touvron2023llama}. However, we argue that the code data is highly important for two reasons. First, the code data is likely to improve the ability of complex reasoning with chain-of-thought~\cite{fu2022gptroadmap}. More importantly, different from natural language, which is sometimes blurry and easy to misunderstand, code is always precise. This enables code to be a more efficient language for machines to convey information concisely without misunderstanding between different (embodied) AI agents, and as a result, code has great potential to dominate LLM communications in real-life applications. Therefore, we design a more code-dominated pre-training data mixture. As shown in Table~\ref{tab:data-mixture}, we extracted 50\% of data from the RedPajama~\cite{together2023redpajama} and 50\% of data from the duplication version of The Stack~\cite{Kocetkov2022TheStack}. Our experimental results show that the version I data mixture might be a bit aggressive in its code proportion. We fix these issues at the later stage of pre-training, please see the following Section~\ref{sec:training_progress} for details.

\begin{table}[h]
\centering
\caption{Three versions of OpenMoE pre-training data mixture.}
\label{tab:data-mixture}
\begin{tabular}{lccc}
\toprule
 & \textbf{Version I} & \textbf{Version II} & \textbf{Version III} \\ 
\textbf{Model} & \multicolumn{2}{c}{\hspace{-1cm}OpenMoE-Base, OpenMoE-8B/32E} & OpenMoE-34B/32E \\
\textbf{Period} & \multicolumn{2}{c}{\hspace{-1cm}before 780B tokens $\rightarrow$ after 780B tokens} & from start to end \\
\midrule
\textbf{Dataset} & \multicolumn{3}{c}{\textbf{Sampling Ratio}}\\
RedPajama & 50.0\% & 83.5\% & 67.5\%\\
\quad C4 & {\small\quad 7.50\%} &  {\small\quad 15.0\%} &  {\small\quad 15.0\%} \\
\quad Wikipedia &  {\small\quad 2.25\%} &  {\small\quad 6.50\%} &  {\small\quad 4.50\%} \\
\quad Stackexchange &  {\small\quad 1.00\%} &  {\small\quad 2.50\%} &  {\small\quad 1.00\%} \\
\quad ArXiv &  {\small\quad 1.25\%} &  {\small\quad 4.50\%} &  {\small\quad 4.50\%} \\
\quad Books &  {\small\quad 2.25\%} &  {\small\quad 6.50\%} &  {\small\quad 4.50\%} \\
\quad GitHub &  {\small\quad 2.25\%} &  {\small\quad 5.00\%} &  {\small\quad 5.00\%} \\
\quad Commoncrawl & {\small\quad 33.5\%} &  {\small\quad 43.5\%} &  {\small\quad 33.0\%} \\
Wikipedia-en & 0.00\% & 6.50\% & 2.50\% \\
The Stack Dedup & 50.0\% & 10.0\% & 30.0\% \\
\bottomrule
\end{tabular}
\end{table}

\subsection{Model Architecture: Decoder-only ST-MoE}

\textbf{Tokenizer.} We applied umT5~\cite{chung2023unimax} tokenizer with 256K vocab size for two reasons: (1) umT5 tokenizer with a large multi-lingual vocab supports low-resource language better than the tokenizers using a small vocab (\eg LLaMA tokenizer with 32K vocab); (2) comparing to some old tokenizers, such as BERT~\cite{kenton2019bert} and T5~\cite{2020t5} tokenizer, umT5 tokenizer has byte fallback feature to support out-of-vocab tokens better.

\textbf{Token-choice Routing.} We generally follow ST-MoE~\cite{zoph2022st} for our model architecture and routing design to ensure training stability, which is extremely important when training larger models. Given $E$ trainable experts and input representation $x\in \mathbb{R}^D$, the output of MoE model can be formulated as:
\begin{equation}
\mathrm{MoE}(x)=\sum_{i=1}^E {g(x)}_i {e_i(x)},
\end{equation}
where ${e_i(\cdot)}$ is a non-linear transformation $\mathbb{R}^D \to \mathbb{R}^D$ of the $i^{\mathrm{th}}$ expert, and ${g(\cdot)}_i$ is the $i^{\mathrm{th}}$ element of the output of the trainable router $g(\cdot)$, a non-linear mapping $\mathbb{R}^D \to \mathbb{R}^E$. Usually, both $e(\cdot)$ and $g(\cdot)$ are parameterized by neural networks. Please note each expert is an FFN layer instead of a complete Transformer model in most MoE-based Transformer models, including ours. 

\textbf{Top-2 Selection.} According to the formulation above, when $g(\cdot)$ is a sparse vector, only part of the experts would be activated and updated by back-propagation during training. We set the gating layer as a top-K selection as:
\begin{equation}\label{eq:topK}
\mathrm{g}(x)=\mathrm{TopK}(\mathrm{softmax}(f(x))),
\end{equation}
where $f(\cdot)$ is routing linear transformation $\mathbb{R}^D \to \mathbb{R}^E$. When $K \ll E$, most elements of $\mathrm{g}(x)$ would be zero so that sparse conditional computation is achieved. We set $K=2$ following \citet{zoph2022st}.

\textbf{Residual MoE.} Each vanilla Transformer block can be written as:
\begin{equation}
\begin{aligned}
x' &= \mathrm{LayerNorm}^{\mathrm{att}}_{i}(x), \\
x &= \mathrm{MHA}(x') + x,\\
x''&= \mathrm{LayerNorm}^{\mathrm{ffn}}_{i}(x), \\
x &= \mathrm{FFN}(x'') + x,\\
\end{aligned}
\end{equation}
In OpenMoE, for each MoE-based Transformer block, we use one residual MoE layer to ensure that one fixed FFN layer is always activated for every token. That is:
\begin{equation}
\begin{aligned}
x' &= \mathrm{LayerNorm}^{\mathrm{att}}_{i}(x), \\
x &= \mathrm{MHA}(x') + x,\\
x''&= \mathrm{LayerNorm}^{\mathrm{ffn}}_{i}(x), \\
x &= \mathrm{MoE}(x'') + \mathrm{FFN}(x'') + x,\\
\end{aligned}
\end{equation}
Note we use MoE-based Transformer blocks in an interleaved manner instead of placing MoE in every Transformer block. In our setting, we use MoE every 4 layers in OpenMoE-Base/16E and OpenMoE 34B/32E and use MoE every 6 layers for OpenMoE-8B/32E. This setting is inspired by the findings in ViT-MoE~\cite{riquelme2021scaling}, \ie using MoE every layer introduces more computational overhead during routing, and then induces a worse cost-effective trade-off than interleaved MoE usage.

\textbf{Load Balance Loss and Router Z-loss.} ST-MoE~\cite{zoph2022st} follows ~\citet{shazeer2017outrageously}, using MoE load balance loss to ensure a balanced number of tokens assigned to different experts so that MoE models can achieve better parallelism. For each routing operation, given $E$ experts and $N$ batches with $B=NL$ tokens, the following auxiliary loss is added to the total model loss during training: 
\begin{equation}\label{eq:balance_loss}
L_{b} = E \cdot \sum_{i=1}^E m_i \cdot P_i,
\end{equation}
where $m$ is a vector, $P_i$ is $\mathrm{softmax}(f(x))$. $i$ denotes the expert ID. The $i^{\mathrm{th}}$ element is the fraction of tokens dispatched to expert $i$:
\begin{equation}
m_i = \frac{1}{B} \sum_{j=1}^{B} \mathrm{h}(x_j)_i,
\end{equation}
where $\mathrm{h}(\cdot)$ is an index vector selected by $\mathrm{TopK}$ in Eq.~\ref{eq:topK}. $\mathrm{h}(x_j)_i$ is the $i^{\mathrm{th}}$ element of $\mathrm{h}(x_j)$. It is noticeable that, different from $g(x)_i$ in Eq.~\ref{eq:topK}, $m_i$ and $\mathrm{h}(x_j)_i$ are non-differentiable. However, a differentiable loss function is required to optimize MoE in an end-to-end fashion, so we use the routing score $\mathrm{softmax}(f(x))$ in Eq.~\ref{eq:topK}  (\ie $P_i$ in Eq.~\ref{eq:balance_loss})to make the routing decision differentiable and then learnable.

In addition to the load balance loss, ~\citet{zoph2022st} proposed router z-loss for more stable MoE training:
\begin{equation}
L_{z}(x) = \frac{1}{B} \sum_{i=1}^{B} \left( \log \sum_{j=1}^{E} e^{x^{(i)}_j} \right)^2
\end{equation}

This router z-loss can penalize large logits input to the gating network and encourage the absolute magnitude of numbers to be small so that it can reduce the round-off errors in MoE layers. Please refer to ST-MoE paper~\cite{zoph2022st} for a detailed explanation.

Taken together, our final training loss can be written as:
\begin{equation}
L = L_{CE} + L_{b} + L_{z}
\end{equation}
where $L_{CE}$ is the cross-entropy loss in language model pre-training.

\subsection{Training Objective: UL2 and CasualLM}

\begin{table}[t]
\centering
\caption{UL2’s mixture-of-denoisers configuration, $\mu$ is average span length and $r$ is the mask ratio.}
\label{tab:ul2_objective_mixture}
\begin{tabular}{l|c}
\toprule
\textbf{Training Objective} & \textbf{Percentage}  \\
\midrule
\textbf{PrefixLM}, $r$=0.5 & 50\%  \\
\textbf{SpanCorrupt} \\
\qquad $\mu$=3, $r$=0.15 & 10\%  \\
\qquad $\mu$=8, $r$=0.15 & 10\% \\
\qquad $\mu$=3, $r$=0.5 & 10\% \\
\qquad $\mu$=8, $r$=0.5 & 10\% \\
\qquad $\mu$=64, $r$=0.5 & 10\% \\
\bottomrule
\end{tabular}
\end{table}

Instead of adopting vanilla casual language modeling (CasualLM) directly, we explore UL2~\cite{tay2022ul2}, a more diverse language model pre-training objective combining span corruption (SpanCorrupt) and prefix language modeling (PrefixLM)~\cite{2020t5}. It is noteworthy that the SpanCorrupt in UL2 is more diverse than the vanilla SpanCorrupt because it mixes various span lengths and corruption rates. We have two reasons to explore UL2 in OpenMoE. First, UL2 has shown promising results in PaLM-2~\cite{anil2023palm}. More importantly, the aggressive token masking is very similar to the code completion task in the real world, such as Copilot. \citet{bavarian2022efficient} also found that the similar filling-in-the-middle (FiM) objective can model the code better than the vanilla training objective. Since we used more code in our pre-training data mixture, adapting UL2 that covers FiM is a more reasonable choice intuitively.

Our detailed UL2 training objective configuration is shown in Table~\ref{tab:ul2_objective_mixture}. We use only 20\% low mask ratio ($r$=0.15) because there are fewer output tokens during training, which may slow down the learning. We also use more PrefixLM than the default UL2 setting because we think the zero-shot and in-context learning ability enhanced by PrefixLM training is important. We faced some difficulties when training with UL2 in OpenMoE, which will be discussed in Section~\ref{sec:training_progress}.

\subsection{Supervised Fine-tuning}

Although alignment is not the focus of this OpenMoE project, we still conduct supervised fine-tuning (SFT) with a subset of the open-sourced WildChat dataset~\cite{anonymous2024inthewildchat} to enhance the instruction following ability and study the behavior of the MoE model before and after SFT. We only pick the instruction-response pairs from GPT-4 in WildChat because of the lack of computation resources at the late stage of OpenMoE development. The subset includes 58K conversations and each conversation includes 1.8 turns on average.

\subsection{Other Designs}

Following recent LLMs, we adopt RoPE~\cite{su2024roformer} for position embedding and SwiGLU~\cite{shazeer2020glu} for activation function for FFNs in both dense and MoE Transformer blocks.
More detailed model configuration and training hyperparameters for OpenMoE models can be found in Appendix~\ref{sec:hyper-param}. We applied data parallelism, tensor parallelism~\cite{xu2021gspmd,shoeybi2019megatron}, and expert parallelism~\cite{lepikhin2020gshard} for training models at scale. We train OpenMoE models on Google Cloud TPU with 64 to 512 v3 chips depending on the availability.

\section{Training OpenMoE}
\label{sec:training}

\begin{table}[t]
\centering
\caption{Ablation study with OpenMoE-Base/16E on zero-shot TriviaQA~\cite{joshi-etal-2017-triviaqa}.}
\label{tab:results}
\begin{tabular}{l|cc}
\toprule
\textbf{Method} & \textbf{EM} & \textbf{F1} \\
\midrule
OpenMoE & 1.4 & 4.5 \\
\quad w/o MoE & 0.1 & 0.3 \\
\quad w/o UL2 (PrefixLM only) & 0.0 & 0.0 \\
\quad w/o Code data & 0.7 & 1.1 \\
\quad w/ LLaMA tokenizer & 2.2 & 5.7 \\
\bottomrule
\end{tabular}
\end{table}

\subsection{Ablation Study} As an initial evaluation of our design decisions, we conducted an ablation study using the OpenMoE-Base/16E model. It's important to note that while these results provide early insights, we cannot be certain of their generalizability to larger models, primarily due to computational resource constraints that preclude larger-scale ablations.

Our findings indicate that several elements — the MoE approach, the UL2 training objective, and the increased emphasis on code data — all contribute positively to the base version's performance in zero-shot TriviaQA tasks. The model using LLaMA tokenizer~\cite{touvron2023llama} outperforms the one with umT5 tokenizer. This outcome is considered acceptable, even though a larger vocabulary size might slightly impair performance. We believe that supporting low-resource languages is crucial, as foundational models should be accessible and beneficial to a diverse global audience. After this sanctity check, we proceed to scale OpenMoE up to OpenMoE-8B/32E.  

\begin{figure}[t]
\centering
\begin{minipage}[t]{0.45\linewidth}
\centering
\includegraphics[width=0.9\textwidth]{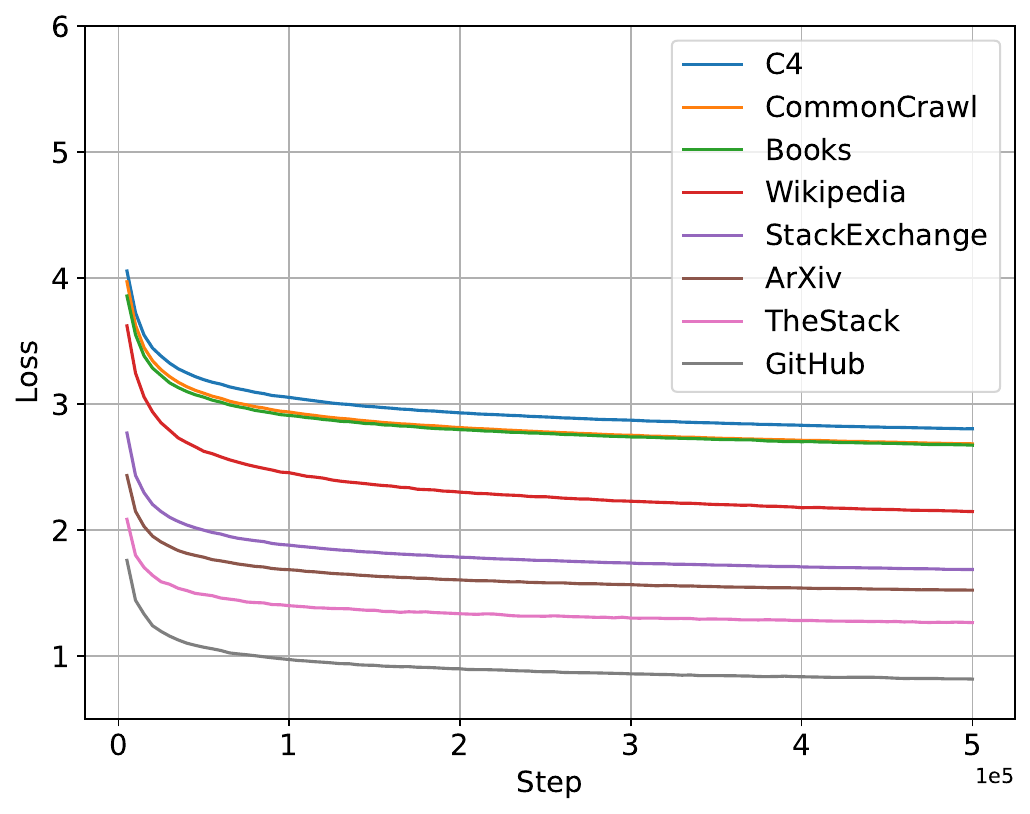}
\subcaption{Comparison of the validation loss on different pre-training datasets.}
\label{fig:moe_base_predict}
\end{minipage}\hspace{0.8cm}
\begin{minipage}[t]{0.45\linewidth}
\includegraphics[width=0.9\textwidth]{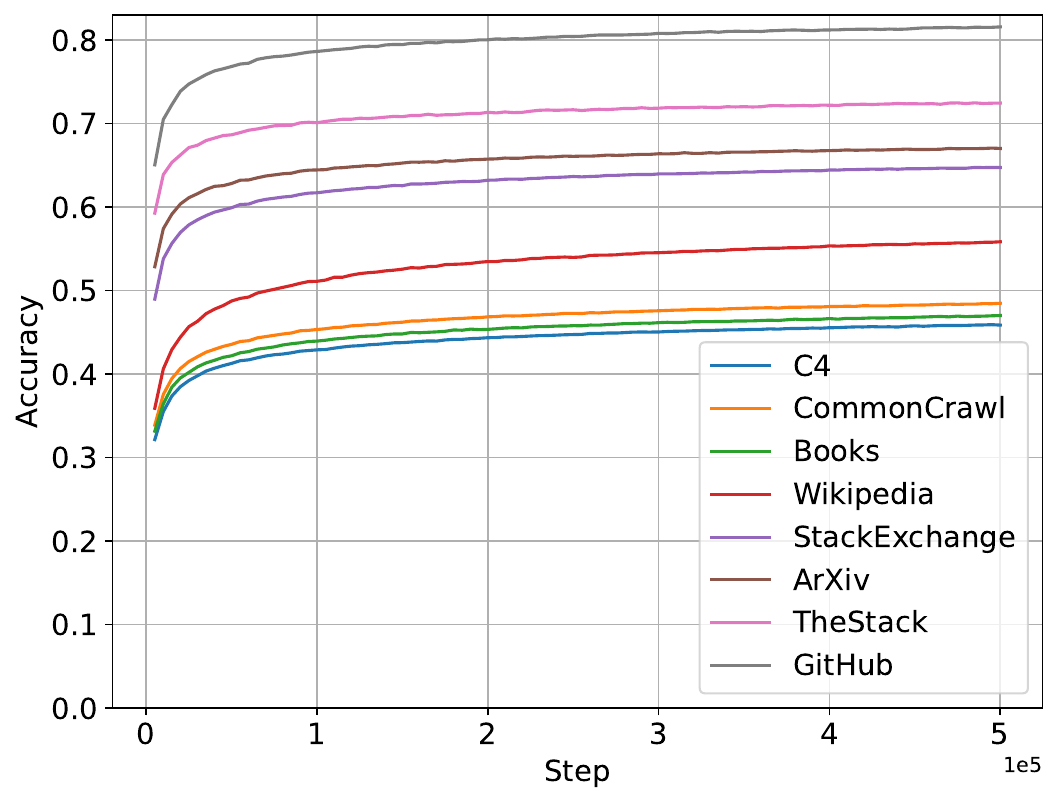}
\subcaption{Comparison of validation accuracy on different pre-training datasets.}
\label{fig:moe_large_predict}
\end{minipage}
\caption{Comparison of the validation loss and accuracy on different pre-training datasets. We can observe that models are easier to achieve higher accuracy and lower loss on code data.}
\label{fig:data_mixture_compare}
\end{figure}

We also conduct an ablation study to compare the progress of learning the data from different domains. As shown in Figure~\ref{fig:data_mixture_compare}, we can observe that models are easier to achieve higher accuracy and lower loss on code data. On Github, although our model is small, it can still achieve over 80\% token prediction accuracy. We infer that this is because of the long-tail token distribution in code data. For instance, a large number of tokens in code are ``\textbackslash n'' and ``\textbackslash t'', which are relatively easier to predict.

\subsection{Training Progress}\label{sec:training_progress}

\begin{figure}[t]
  \begin{center}
    \includegraphics[width=0.45\textwidth]{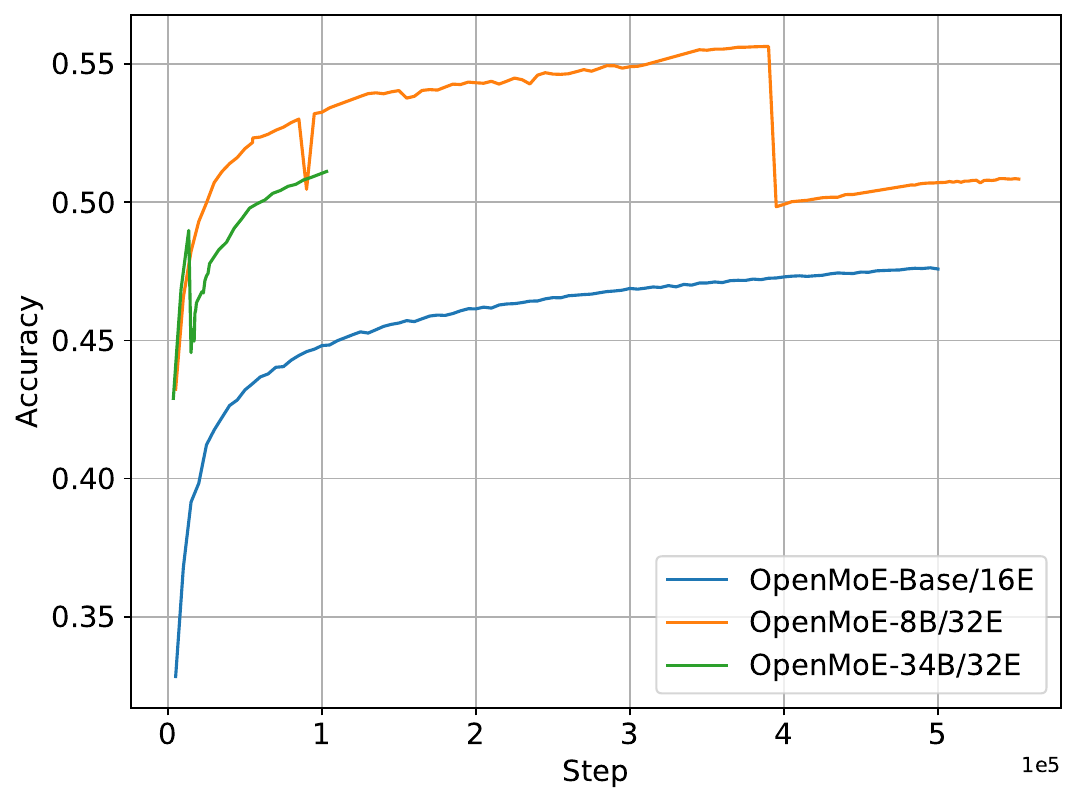}
  \end{center}
  \caption{Token prediction accuracy of OpenMoE models. OpenMoE-8B/32E uses UL2 before 390K steps and falls back to CasualLM after 790K steps. OpenMoE-34B/32E uses UL2 until 50B tokens.}\label{fig:openmoe_models}
\end{figure}


\textbf{UL2 Saturation} During training, we found that, although UL2 can help the model to learn faster at the early stage of training, it is easier to saturate at the later training stage of OpenMoE-8B/32E. As shown in Figure~\ref{fig:openmoe_models}, if we zoom in, we can find that OpenMoE-8B/32E improves very slowly from 35K to 39K steps. We suggest that this may be because, although UL2 is more diverse, the SpanCorrupt is still relatively easy compared to CasualLM. Therefore, we fall back to CasualLM after 390K steps (780B) tokens. In addition, since code data aligns better with UL2 and our initial code data mixture is relatively aggressive, we also decreased our code data sampling ratio to 15\%. The Second version data mixture is reported in Table~\ref{tab:data-mixture}.



Obviously, in Figure~\ref{fig:openmoe_models}, after 780B tokens, there is a significant drop in the token prediction accuracy after 390K steps for OpenMoE-8B/32E. This is caused by the more difficult CasualLM objective and less easy code data. Note that, although we encountered a saturation issue at the later stage of OpenMoE-8B/32E training, we think such an easy-to-hard curriculum may be helpful for LLM training. Therefore, we still adapted UL2 for 25K steps (50B tokens) in OpenMoE-34B/32E. We used a relatively moderate code-heavy data mixture in OpenMoE-34B/32E. As shown in Table~\ref{tab:data-mixture}, we utilize 35\% of code data in total. Due to the computation resource limitation, we train OpenMoE-34B/32E with only 200B tokens to verify its scalability. We leave training a large-scale OpenMoE with more tokens as future work if possible.

\subsection{Evaluation on Benchmarks}
\label{sec:eval}

\subsubsection{Raw Model Evaluation}

Before all, we highlight that we did not hack the benchmarks at all and the pre-training is purely on the open-sourced datasets mentioned above. Since our model is relatively small in terms of training budget, we mainly evaluate the raw model on established but not that hard benchmarks, \ie TriviaQA~\cite{joshi-etal-2017-triviaqa}, HumanEval~\cite{chen2021codex}, WMT16-En-Ro~\cite{bojar-EtAl:2016:WMT1}, BigBench-Lite (24 tasks)~\cite{srivastava2023beyond}, and a subset of the lm-evaluation-harness collection~\cite{eval-harness} with 13 tasks. For popular but relatively challenging benchmarks like 5-shot MMLU~\cite{hendrycks2020measuring}, our OpenMoE-8B/32E achieves around 26.2\% accuracy, which means the model is almost randomly guessing from the four options. We mainly compare with the open-sourced models with more training cost, \ie TinyLLaMA-1.1B~\cite{zhang2024tinyllama} and OpenLLaMA-3B~\cite{openlm2023openllama}. On BigBench-Lite, we also compare with GPT-3~\cite{brown2020language}, Big-G~\cite{srivastava2023beyond} and Big-G-Sparse~\cite{srivastava2023beyond}. Big-G and Big-G-Sparse are two sets of Google in-house Transformer models evaluated on BigBench-Lite, and Big-G-Sparse models are MoE-based Transformers.

\begin{table}[t]
\centering
\caption{Results on TriviaQA (Exact Match). We also report the number of training tokens from Wikipedia because the commonsense questions in TriviaQA have a relatively close relation with Wikipedia data.}
\label{tab:results-triviaqa}
\begin{tabular}{lccccc}
\toprule
\textbf{Model} & \textbf{Act. Params} & \textbf{Total Tokens} & \textbf{Text Tokens} & \textbf{Wiki Tokens} & \textbf{TriviaQA} \\
\midrule
TinyLLaMA-1.1B & 0.9B & 3.0T & 2.1T & 75B & 11.2 \\
OpenLLaMA-3B & 2.9B & 1.0T & 991B & 24B & 29.7 \\
OpenMoE-8B/32E & 2.1B & 1.1T & 644B & 58B & 32.7 \\
OpenMoE-34B/32E & 6.4B & 0.2T & 130B & 14B & 31.3 \\
\bottomrule
\end{tabular}
\end{table}

We first report our results on Commonsense QA (TriviaQA), Coding (HumanEval), and Low-resource Machine Translation (WMT16 En-Ro). We think these three benchmarks are meaningful for us because (1) Commonsense is to check whether OpenMoE can memorize more commonsense given its efficient parameter scaling advantage; (2) Coding is important because of its prevalent use cases in solving coding-related user prompts, LLM as agents, and embodied AI; (3) Low-resource Machine Translation is important because we want to share the benefits of foundation models to everyone on earth. As shown in Table~\ref{tab:results-triviaqa}, OpenMoE-8B/32E outperforms baselines clearly with less training cost (Activated parameter$\times$Total Training tokens). Also, please note that TinyLLaMA-1.1B performs significantly worse than other models on TriviaQA although it has a comparable training cost with OpenLLaMA-3B. Therefore, this highlights the importance of the number of parameters to keeping knowledge in LLMs, which also indicates the significance of using MoE.

\begin{table}[t]
\centering
\caption{Results on HumanEval (Pass@1). We also report the number of training tokens from the code domain (The Stack and GitHub data).}
\label{tab:results-humaneval}
\begin{tabular}{lcccc}
\toprule
\textbf{Model} & \textbf{Act. Params} & \textbf{Total Tokens} & \textbf{Code Tokens} & \textbf{HumanEval} \\
\midrule
TinyLLaMA-1.1B & 0.9B & 3.0T & 900B & 9.1 \\
OpenLLaMA-3B & 2.9B & 1.0T & 59B & 0 \\
OpenMoE-8B/32E & 2.1B & 1.1T & 456B & 9.8 \\
OpenMoE-34B/32E & 6.4B & 0.2T & 70B & 10.3 \\
\bottomrule
\end{tabular}
\end{table}

In Table~\ref{tab:results-humaneval}, OpenMoE models achieve better performance than baselines. OpenMoE-34B/32E only used 70B code data, while it still performs relatively well on HumanEval, which shows the scalability of OpenMoE, although we don't have enough computation resources to train it until the end. OpenLLaMA-3B struggles on HumanEval because consecutive whitespaces are treated as one, contradicting the Python syntax\cite{nijkamp2023xgen}.

\begin{table}[t]
\centering
\caption{Results on WMT16 En-Ro (BLEU score). We also report the number of explicit multi-lingual tokens in the pre-training dataset, \ie the multi-lingual version of Wikipedia from the RedPajama dataset.}
\label{tab:results-mt}
\begin{tabular}{lccccc}
\toprule
\textbf{Model} & \textbf{Act. Params} & \textbf{Total Tokens}  & \textbf{Multi-lingual Tokens} & \textbf{WMT16 En-Ro} \\
\midrule
TinyLLaMA-1.1B & 0.9B & 3.0T & 75B & 2.6 \\
OpenLLaMA-3B & 2.9B & 1.0T & 24B & 1.9 \\
OpenMoE-8B/32E & 2.1B & 1.1T & 38B & 3.1 \\
OpenMoE-34B/32E & 6.4B & 0.2T & 9B & 3.4 \\
\bottomrule
\end{tabular}
\end{table}

Table~\ref{tab:results-mt} shows our results on WMT16 En-Ro translation task. Note that our model did not include much multi-lingual data intentionally. Most multi-lingual data should be from the multi-lingual version of Wikipedia in the RedPajama, which is also used in TinyLLaMA-1.1B and OpenLLaMA-3B. However, OpenMoE models still show better results than baselines, which potentially highlights the importance of umT5 tokenizer.


\begin{figure}
  \begin{center}
    \includegraphics[width=0.8\textwidth]{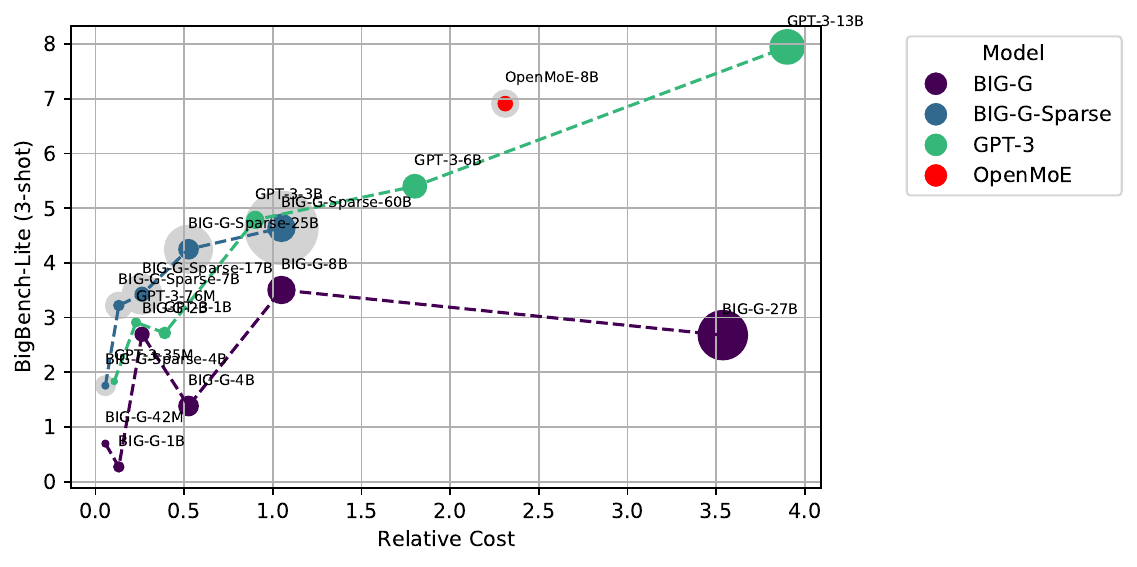}
  \end{center}
  \caption{Results on BigBench-Lite. The relative cost is computed based on multiplying activated parameters in the Transformer and the number of training tokens. The size of the color dots denotes the number of activated parameters, and the size of the shadow denotes the number of total parameters for MoE models.}\label{fig:bblite-results}
\end{figure}

In Figure~\ref{fig:bblite-results}, the relative cost is computed based on multiplying activated parameters (Act. Params) in Transformer blocks and the number of training tokens. The size of the color dots denotes the number of activated parameters, and the size of the shadow denotes the number of total parameters for MoE models. We can observe that OpenMoE achieved a better cost-effectiveness trade-off on BigBench-Lite, in terms of both training and inference cost.

\begin{table}[t]
\centering
\caption{Evaluate OpenMoE-8B/32E on lm-evaluation-harness. The results of OpenLLaMA are from its homepage, which only provides two effective digits.}
\label{tab:lm-evaluation-harness}
\begin{tabular}{lccc}
\toprule
\textbf{Dataset} & \textbf{TinyLLaMA-1.1B} & \textbf{OpenLLaMA-3B} & \textbf{OpenMoE-8B/32E} \\
\midrule
ANLI-R1 & 34.2 & 33.0 & 32.7 \\
ANLI-R2 & 32.4 & 36.0 & 33.2 \\
ANLI-R3 & 35.1 & 38.0 & 33.9 \\
HellaSwag & 59.2 & 52.0 & 45.5 \\
WinoGrande & 59.1 & 63.0 & 60.3 \\
PIQA & 73.3 & 77.0 & 74.2 \\
ARC-Easy & 55.2 & 68.0 & 64.1 \\
ARC-Challenge & 30.1 & 34.0 & 30.3 \\
Boolq & 57.8 & 66.0 & 61.2 \\
TruthfulQA & 37.6 & 35.0 & 36.0 \\
OpenbookQA & 21.8 & 26.0 & 24.6 \\
RTE & 51.9 & 55.0 & 53.4 \\
WiC & 50.1 & 50.0 & 49.8 \\
\midrule
Average & 45.9 & \textbf{48.7} & 46.1 \\
\bottomrule
\end{tabular}
\end{table}

We also evaluate OpenMoE on the 13 tasks from the LM-Evaluation-Harness collection. As shown in Table~\ref{tab:lm-evaluation-harness}, both OpenMoE and TinyLLaMA performed worse than OpenLLaMA. However, the scores achieved by OpenMOE are acceptable. We suggest that the initial high sampling rate on the code data may harm the results on these text-dominated benchmarks, which is one of the issues we will discuss in Section \ref{sec:rethink}.


\subsubsection{Chat Model Evaluation}

\begin{figure}
\centering
\begin{minipage}[b]{0.49\linewidth}
\centering
\includegraphics[width=1.0\textwidth]{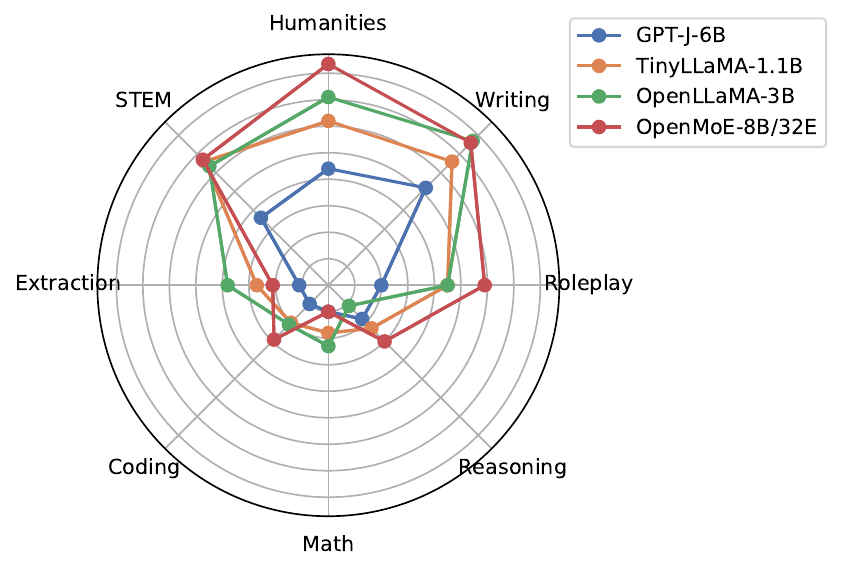}
\subcaption{Single-turn results.}\label{fig:mtbench_turn_1}
\end{minipage}\hspace{0.1cm}
\begin{minipage}[b]{0.49\linewidth}
\includegraphics[width=1.0\textwidth]{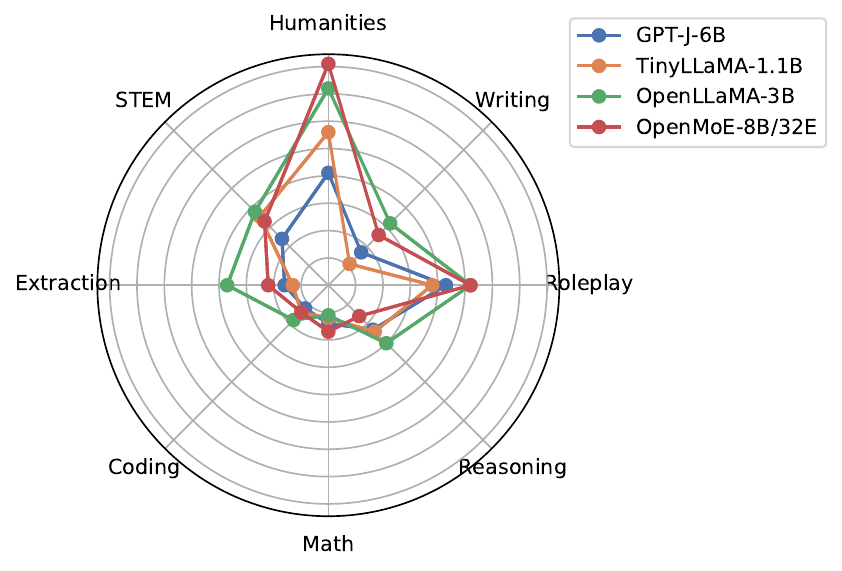}
\subcaption{Multi-turn results.}\label{fig:mtbench_turn_0}
\end{minipage}
\caption{Evaluate OpenMoE on MTBench.}
\end{figure}

\begin{table}[t]
\centering
\caption{Average scores on MT-Bench.}
\label{tab:mtbench_table}
\begin{tabular}{l|ccc}
\toprule
Model & MT-Bench 1st Turn & MT-Bench 2nd Turn & MT-Bench Avg \\
\midrule
GPT-J-6B (0.4T) & 2.51 & 2.35 & 2.43 \\
TinyLLaMA-1.1B (3T) & 4.08 & 2.54 & 3.31 \\
OpenLLaMA-3B (1T) & 4.36 & \textbf{3.62} & \textbf{3.99} \\
OpenMoE-8B/32E (1.1T) & \textbf{4.69} & 3.26 & \textbf{3.98} \\
\bottomrule
\end{tabular}
\end{table}

We further evaluate our model on MTBench, an established ChatBot benchmark that is able to examine models comprehensively. We report both single-turn and multi-turn results in Figure~\ref{fig:mtbench_turn_1} and Table~\ref{tab:mtbench_table}. We can observe that OpenMoE outperforms baselines by a large margin on the single-turn results, especially on coding tasks. However, OpenMoE's performance drops more on the second turn, which results in worse multi-turn results in Figure~\ref{fig:mtbench_turn_0}. We found that this probably be caused by the token drop of a long sequence. Please see the following Section~\ref{sec:analysis} for a detailed analysis.

\section{Analyzing OpenMoE}\label{sec:analysis}

We generally think MoE is an effective way to scale parameters up with a fixed computation budget. However, we have little idea about what the experts in MoE specialize in. In this section, we conduct an in-depth analysis of OpenMoE in multiple aspects to study the routing behavior. 

\subsection{What are the Experts Specializing in?}

\begin{figure}
  \begin{center}
    \includegraphics[width=0.6\textwidth]{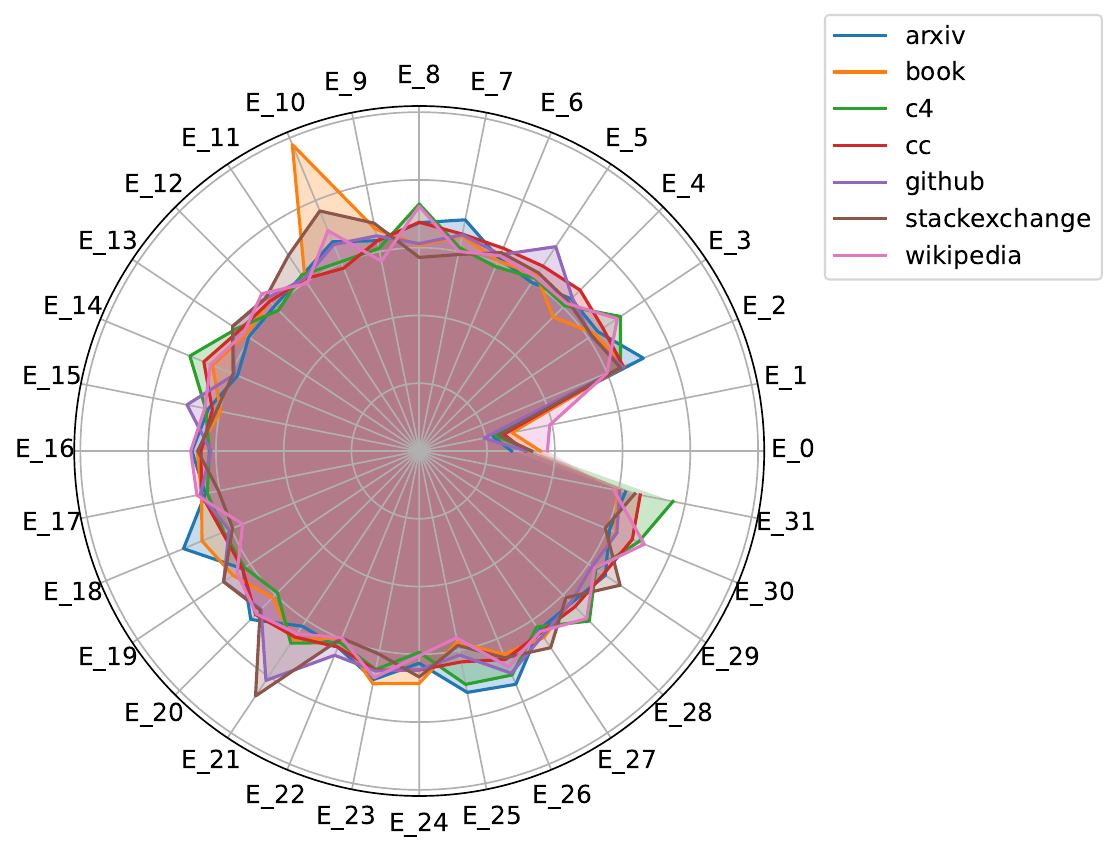}
  \end{center}
  \caption{Visualization of the routing decision on the RedPajama dataset. $E_i$ denotes the ratio of tokens routed to $i_{\mathrm{th}}$ expert.}\label{fig:expert_specialize_redpajama}
\end{figure}

\textbf{Does MoE specialize in domain level?} We first visualize the routing decision of the tokens from different subsets in the RedPajama dataset. Note that all visualization results are from the third MoE layer by default because we did not observe significant differences across layers. We can observe that the tokens from different subsets (\ie domains) are uniformed distributed on the plot. That is, although $E_{21}$ slightly prefers code tokens and $E_{10}$ like books a little, most experts in MoE are not specialized based on the domains.

\begin{figure}[t]
  \begin{center}
    \includegraphics[width=0.55\textwidth]{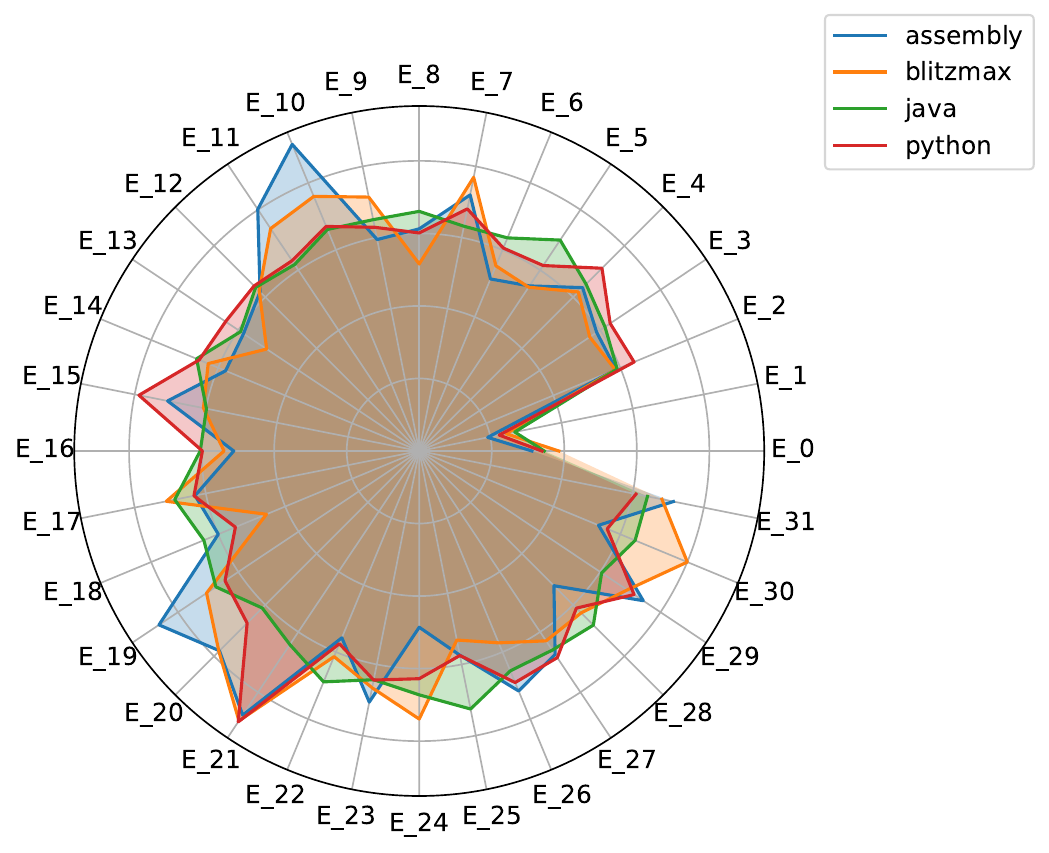}
  \end{center}
  \caption{Visualization of the routing decision on TheStack dataset. $E_i$ denotes the ratio of tokens routed to $i_{\mathrm{th}}$ expert.}\label{fig:expert_specialize_thestack}
\end{figure}

\textbf{Does MoE specialize in language level?} We move forward toward finer-grain data to check whether MoE specializes in different coding languages and natural languages. In Figure~\ref{fig:expert_specialize_thestack}, we compare 4 different coding languages, \ie Assembly, Blitzmax, Java, and Python. Similar to the domain level, even for Assembly and Blitzmax, \ie two low-resource languages compared with Java and Python, they still did not exhibit significant expert specialization.

\begin{figure}
  \begin{center}
    \includegraphics[width=0.5\textwidth]{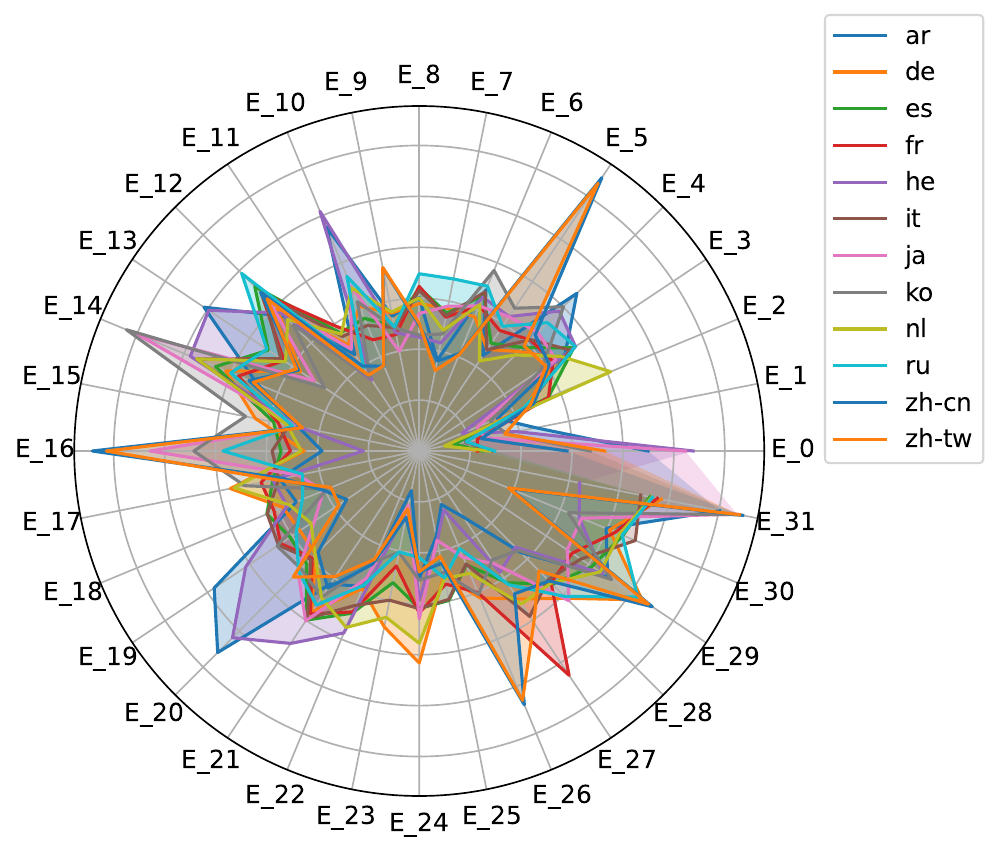}
  \end{center}
  \caption{Visualization of the routing decision on TED-Parallel-Corpus including 12 languages, \ie ar (Arabic), de (German), es (Spanish), fr (French), he (Hebrew), it (Italian), ja (Japanese), ko (Korean), nl (Dutch), ru (Russian), zh-cn (Chinese Simplified), zh-tw (Chinese, Traditional), $E_i$ denotes the ratio of tokens routed to the $i_{\mathrm{th}}$ expert.}\label{fig:expert_specialize_tedmultilingual}
\end{figure}

We further study the expert specialization on different natural languages. We adopted a multi-lingual parallel corpus, \ie TED-Parallel-Corpus~\footnote{https://github.com/ajinkyakulkarni14/TED-Multilingual-Parallel-Corpus} as the platform. In Figure~\ref{fig:expert_specialize_tedmultilingual}, we found that there is a relatively clear specialization among different experts. For instance, zh-cn (Chinese, Simplified) and zh-tw (Chinese, Traditional) both have a strong preference for $E_5$ and $E_{16}$; ja (Japanese), and ko (Korean) both prefer $E_{14}$.

\begin{figure}
  \begin{center}
    \includegraphics[width=0.55\textwidth]{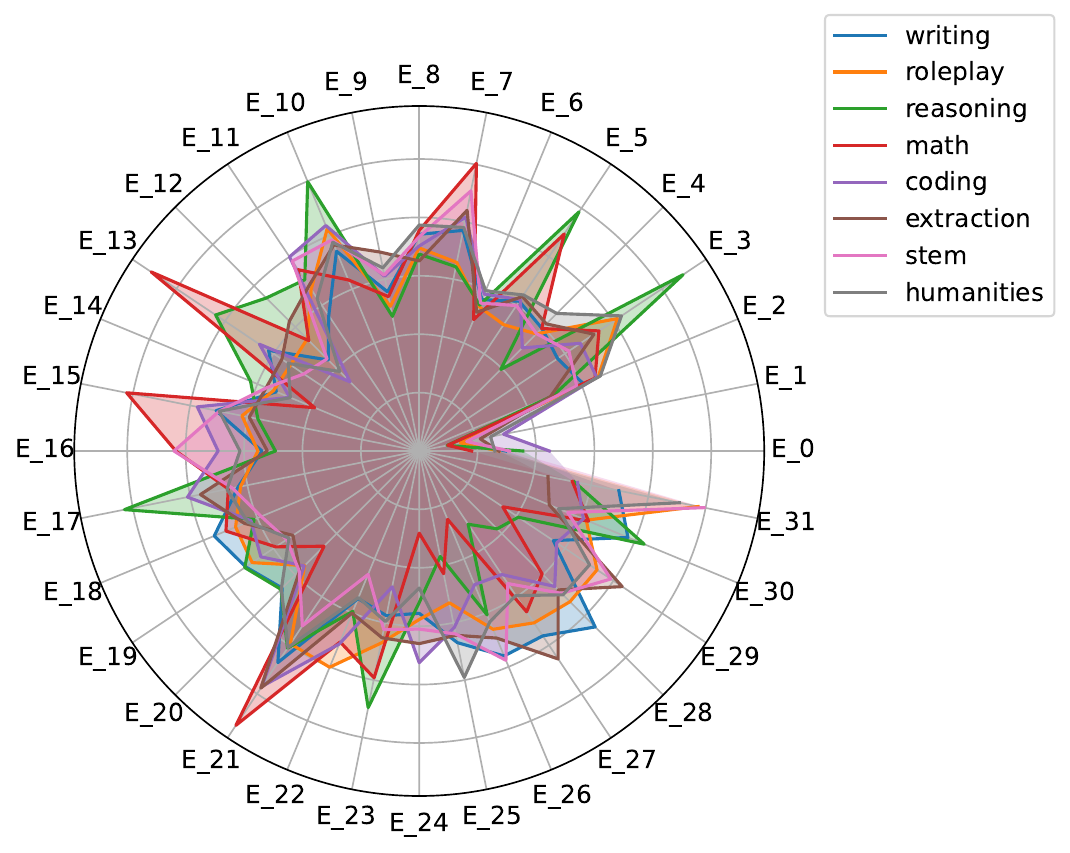}
  \end{center}
  \caption{Visualization of the routing decision on MT-Bench. We adopt the conversation history when evaluating OpenMoE MT-Bench as the visualization data source. $E_i$ denotes the ratio of tokens routed to the $i_{\mathrm{th}}$ expert.}\label{fig:expert_specialize_mtbench}
\end{figure}

\textbf{Does MoE specialize in task level?} Based on the findings above, finer-grained data has clearer expert specialization observation. We then visualize the routing decision on MT-Bench conversation data in Figure~\ref{fig:expert_specialize_mtbench}. We can see a similar specialization as above, especially for the math data. We suggest that the main reason is that the math tasks include more special tokens than other tasks.

\begin{figure}
\begin{minipage}[b]{0.48\linewidth}
  \begin{center}
    \includegraphics[width=1.0\textwidth]{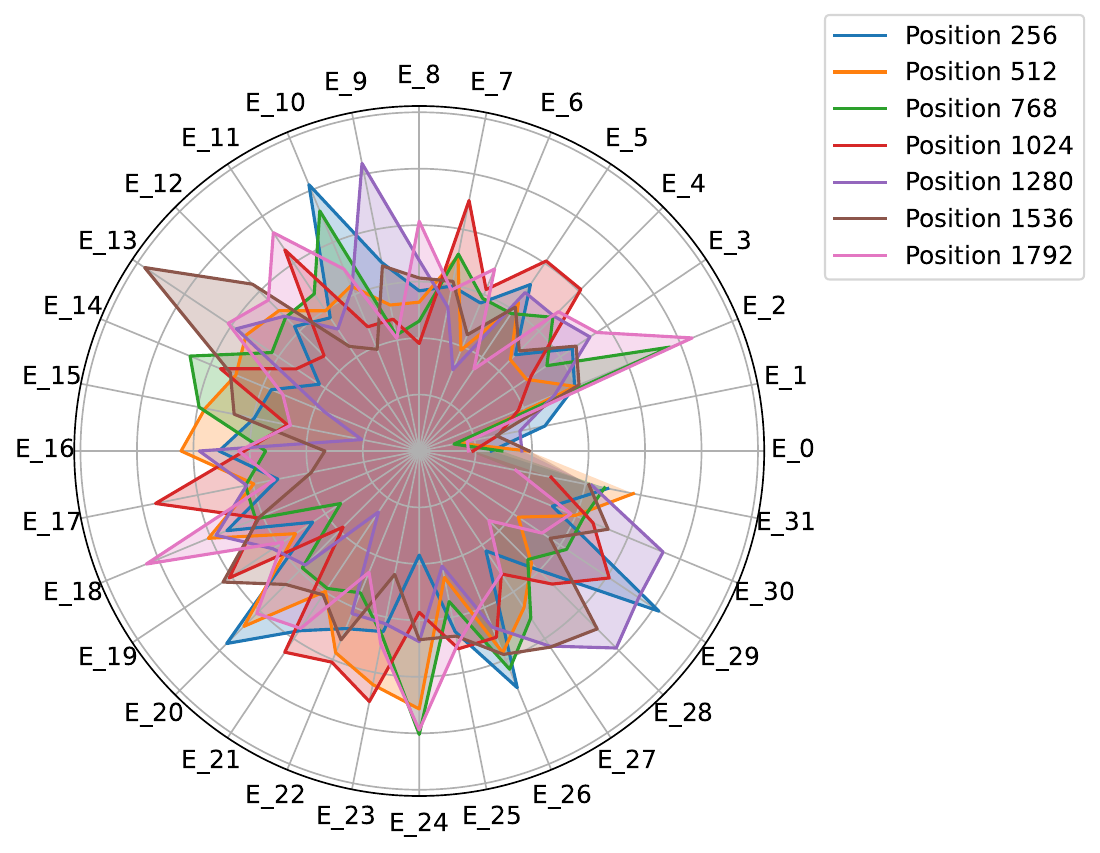}
  \end{center}
  \subcaption{
Uniform sampled token IDs. }\label{fig:expert_specialize_postion}
\end{minipage}\hspace{0.3cm}
\begin{minipage}[b]{0.48\linewidth}
  \begin{center}
    \includegraphics[width=1.0\textwidth]{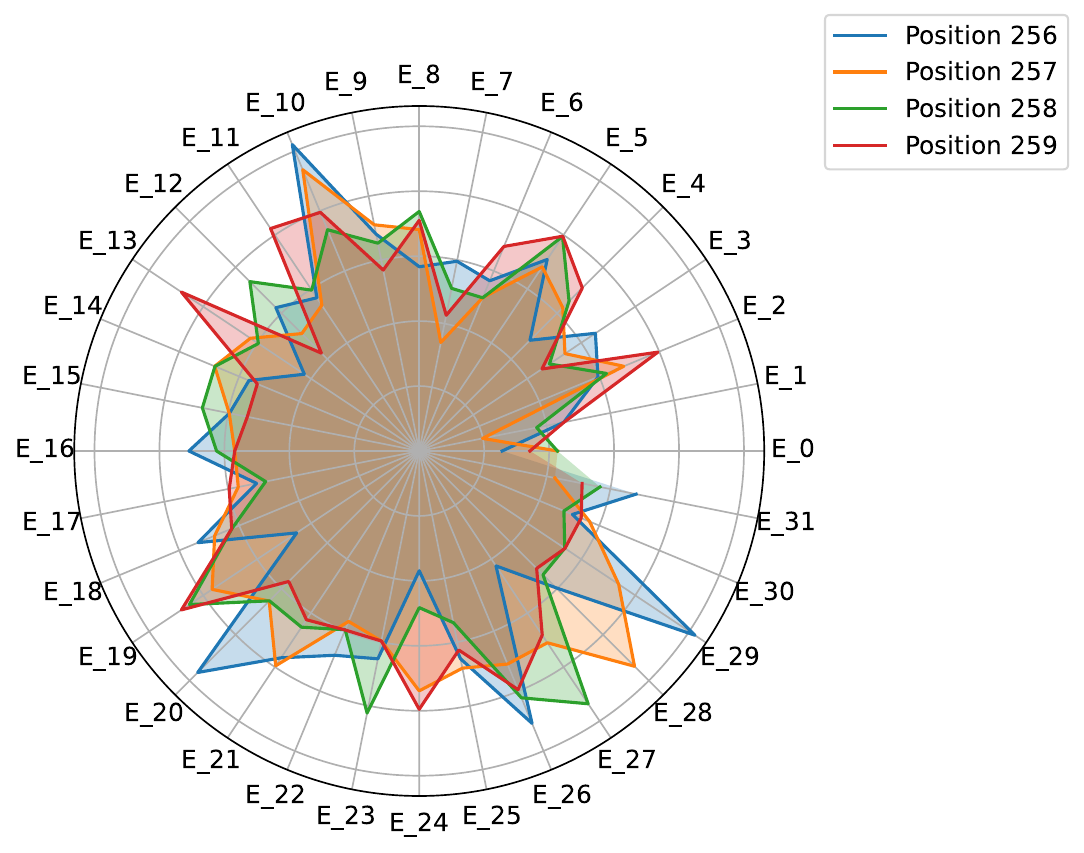}
  \end{center}
  \subcaption{Consecutive token IDs.}\label{fig:expert_specialize_postion_continue}
\end{minipage}
\caption{Visualization of the routing decision at different Position IDs. $E_i$ denotes the ratio of tokens routed to the $i_{\mathrm{th}}$ expert.}
\end{figure}

\textbf{Does MoE specialize in Position ID?} Routers in MoE make decisions based on the token representations. The token representations are from token embeddings and position embeddings. We thus visualize the routing decisions on different positions in Figure~\ref{fig:expert_specialize_postion} and Figure~\ref{fig:expert_specialize_postion_continue}. We can observe:(1) there are indeed some specializations in different 
Position IDs; (2) consecutive positions prefer similar experts, such as the $E_{10}$ and $E_{19}$ in Figure~\ref{fig:expert_specialize_postion_continue}.

\begin{figure}
  \begin{center}
    \includegraphics[width=0.45\textwidth]{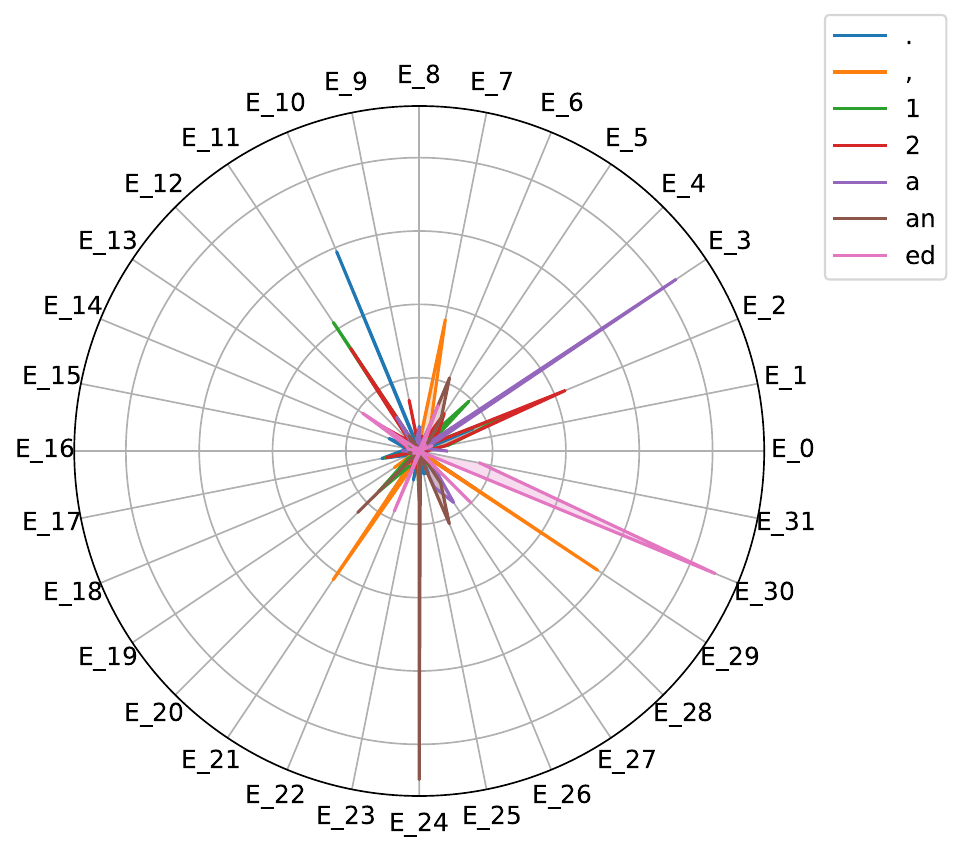}
  \end{center}
  \caption{Visualization of the routing decision at different Token IDs. $E_i$ denotes the ratio of tokens routed to the $i_{\mathrm{th}}$ expert.}\label{fig:expert_specialize_token_id}
\end{figure}

\textbf{Does MoE specialize in Token ID?} Since we are using the umT5 tokenizer, tokens from different languages usually have different token IDs. Therefore, we further study whether the router in MoE mainly makes its decisions based on the Token ID. We visualize the routing decisions of a few representative tokens in Figure~\ref{fig:expert_specialize_token_id}. All these tokens show a very strong specialization on only a few experts. This is a very interesting finding because the tokens with the same Token ID have very diverse contexts in different sentences. For instance, the token ``ed'' can be the suffix of many different words, \eg ``preferred'', and ``led''. The token ``an'' can also be part of ``an apple'' or "another". However, all these tokens have very strong specialization on only a few fixed experts. That means, MoE simply routes based on the Token ID instead of high-level semantics. We name this observation as \textbf{Context-independent Specialization} in the following sections. To verify that the Context-independent Specialization also exists for other Token IDs, we plot the routing decision standard deviation in Appendix~\ref{sec:routing_std}.

\subsection{Token Specialization Study}

\begin{table}[t]
\centering
\caption{Top Tokens selected by each expert.}
\label{tab:top_token_table}
\begin{tabular}{c|l}
\toprule
Expert ID & Top Tokens \\
\midrule
0 & \tokens{\textbackslash n}, \tokens{`}, \tokens{’}, \tokens{s}, \tokens{-}, \tokens{\$}, \tokens{y}, \tokens{\_}, \tokens{\,}, \tokens{2} \\
1 & \tokens{\textbackslash n}, \tokens{1}, \tokens{\,}, \tokens{2},  \tokens{\textbackslash\textbackslash}, \tokens{S}, \tokens{.}, \tokens{-}, \tokens{C}, \tokens{\{} \\
21 & \tokens{,}, \tokens{and}, \tokens{\,}, \tokens{.}, \tokens{\textbackslash n}, \tokens{=}, \tokens{\textbackslash t}, \tokens{the}, \tokens{\,}, \tokens{n} \\
30 & \tokens{\}}, \tokens{ed}, \tokens{d}, \tokens{have}, \tokens{ing}, \tokens{,}, \tokens{has}, \tokens{s},\tokens{"}, \tokens{had} \\
31 & \tokens{to}, \tokens{can}, \tokens{s}, \tokens{of}, \tokens{ing}, \tokens{will}, \tokens{not}, \tokens{e}, \tokens{ed}, \tokens{would} \\
\bottomrule
\end{tabular}
\end{table}

\textbf{Are experts clustering similar tokens?} As we discussed above, the tokens with the same Token ID are always routed to the same expert no matter what the context is, \ie Context-independent Specialization. We thus investigate whether the experts prefer the Token IDs corresponding to the tokens with similar low-level semantics. We list the top 10 favorite tokens for each expert in Table~\ref{tab:top_token_table}. We can observe that similar tokens are clustered in experts. For instance, ``can''. ``will'', and ``would'' are all in expert 31. ``have''. ``has'', and ``had'' are all included in expert 30. This visualization can also explain many observations above. An example is that, in most figures above, we can find most coding and math data prefer expert 21. Here it reveals the real reason. Expert 21 has a strong preference for ``='', ``and'', and ``\textbackslash n'', which appear more frequently in math and code.



\begin{figure}
\begin{minipage}[b]{0.48\linewidth}
  \begin{center}
    \includegraphics[width=1.0\textwidth]{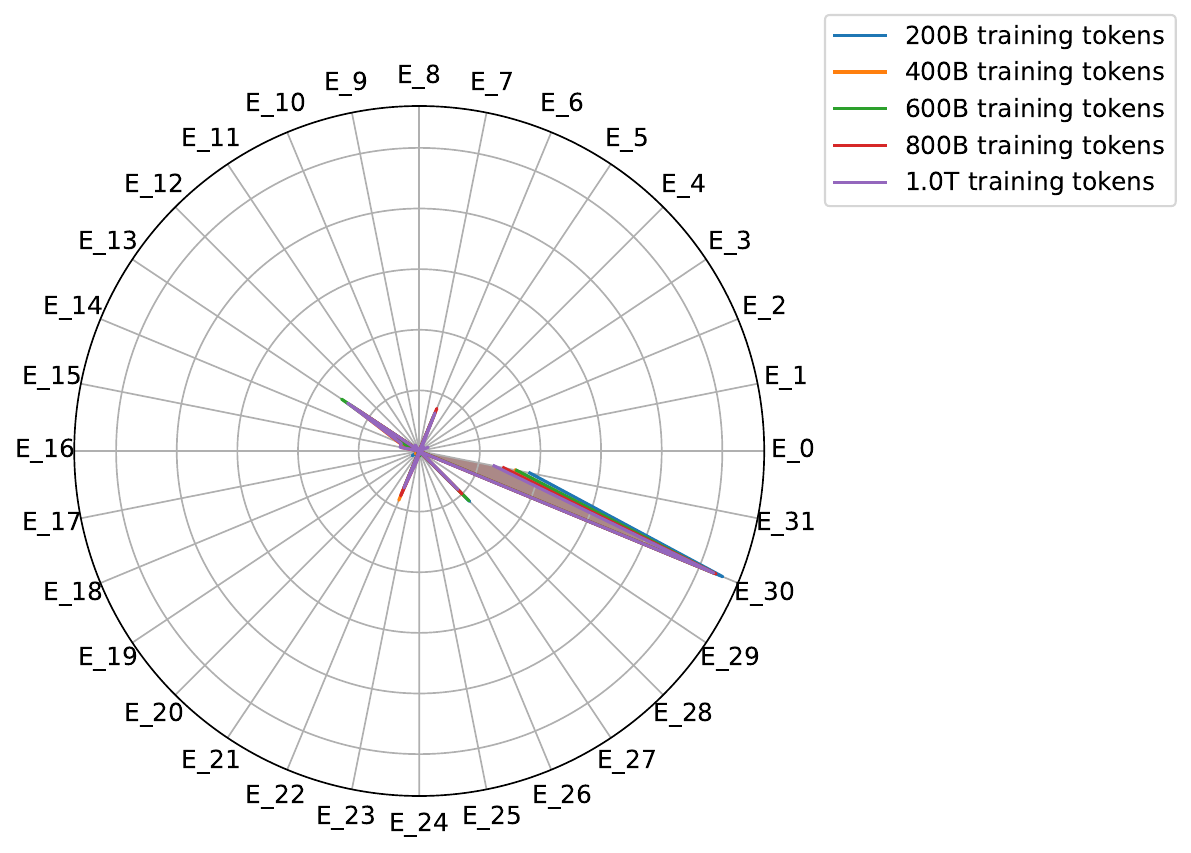}
  \end{center}
  \subcaption{Token ``ed'' routing decision of different intermediate checkpoints.}\label{fig:expert_specialize_token_ed_ckpt}
\end{minipage}\hspace{0.3cm}
\begin{minipage}[b]{0.48\linewidth}
  \begin{center}
    \includegraphics[width=1.0\textwidth]{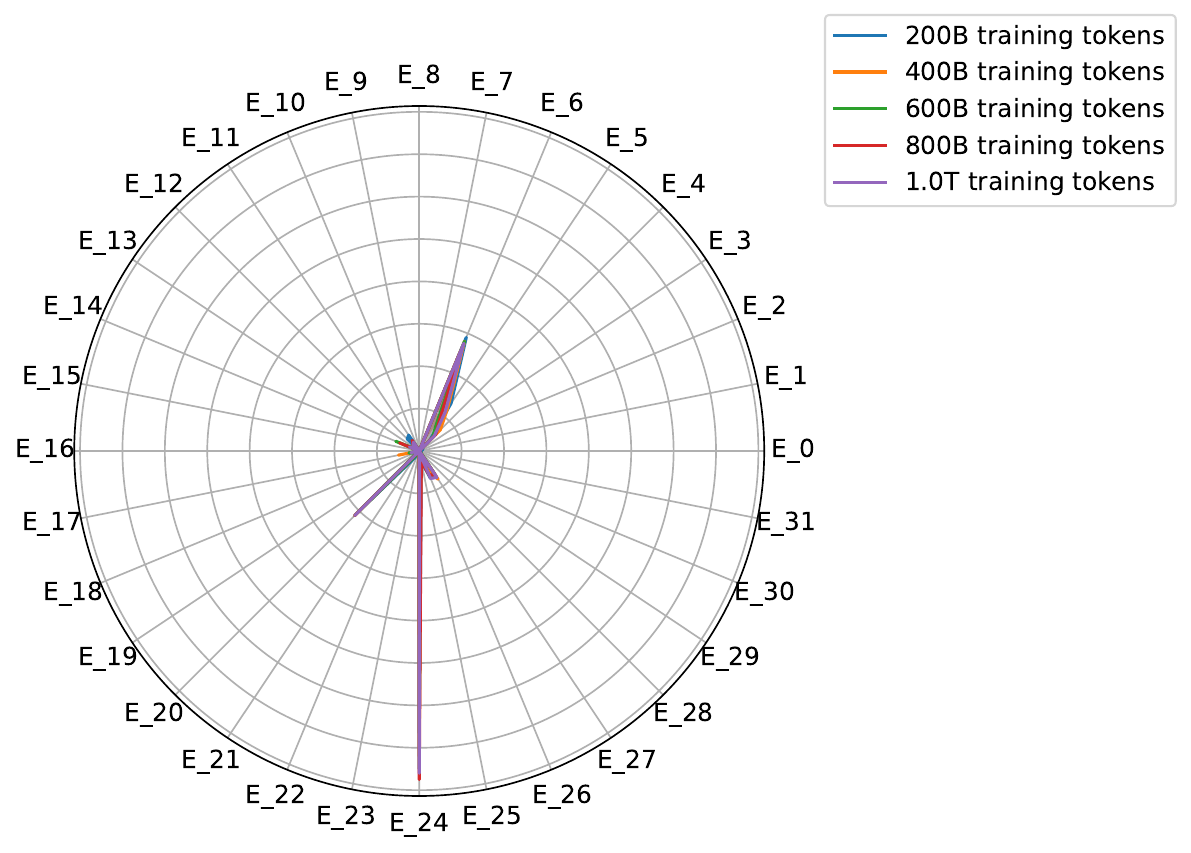}
  \end{center}
  \subcaption{Token ``an'' routing decision of different intermediate checkpoints.}\label{fig:expert_specialize_token_an_ckpt}
    
\end{minipage}
\caption{Visualization of token IDs' routing decision of different intermediate checkpoints. $E_i$ denotes the ratio of tokens routed to the $i_{\mathrm{th}}$}
\end{figure}

\textbf{When did the model learn the specialization?} According to the Context-independent Specialization observed above, the model is not learning how to route based on high-level semantics. Therefore, we raise another question, when did the model learn and fix the routing decision for the tokens? We compare the routing decisions of different OpenMoE intermediate checkpoints in Figure~\ref{fig:expert_specialize_token_ed_ckpt} and Figure~\ref{fig:expert_specialize_token_an_ckpt}. We can see that the expert preferences are almost totally overlapped for different checkpoints, which means that the model has started to fix its routing at the very early stage of training. Even if we change the training data mixture (from 52.25\% code to 20\% code) and training objective (from UL2 to CasualLM), the routing decision is still fixed. We infer that the reason is that, when the token is usually assigned to one specific expert, the loss would increase a lot if the token is sent to another unseen expert, which pushes the model to assign the token back to the original expert. Therefore, the routing probably has been learned at the warmup stage or so, and kept throughout the whole following training stage.

\subsection{Token Drop During Routing}

\begin{figure}
\begin{minipage}[b]{0.48\linewidth}
  \begin{center}
    \includegraphics[width=1.0\textwidth]{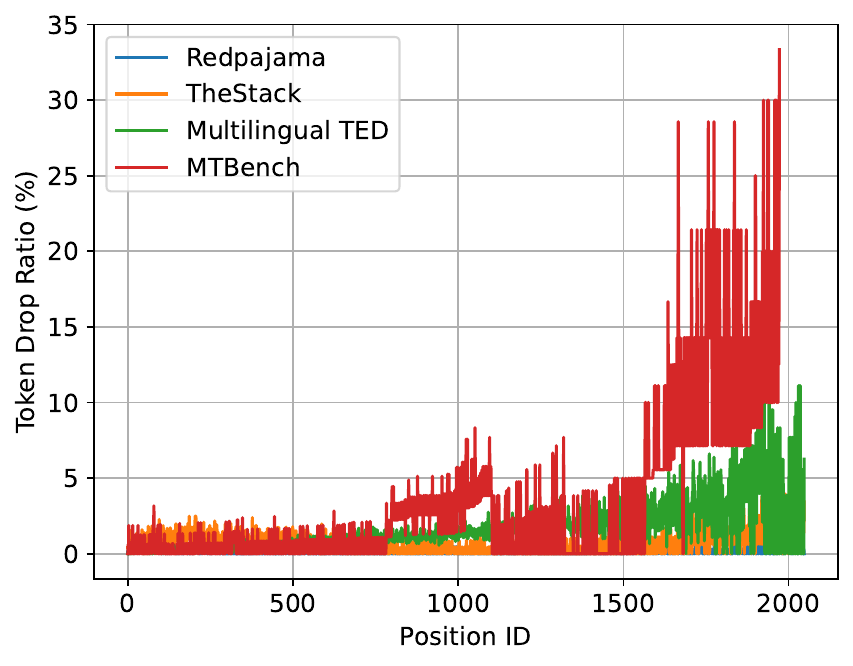}
  \end{center}
  \subcaption{Different datasets.}\label{fig:position_drop_dataset}
\end{minipage}
\begin{minipage}[b]{0.48\linewidth}
  \begin{center}
    \includegraphics[width=1.0\textwidth]{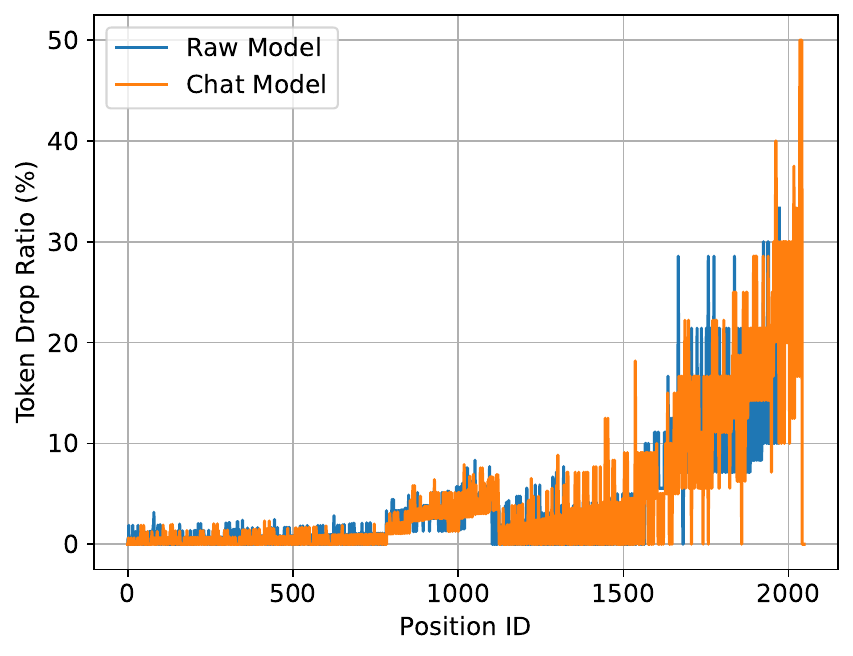}
  \end{center}
  \subcaption{Before and after supervised fine-tuning.}\label{fig:position_drop}
\end{minipage}
\caption{Comparing the ratio of tokens dropped at different position IDs.}
\end{figure}

\textbf{Drop-towards-the-End} In MoE models, we usually set a pre-defined max capacity $C$ for every expert to ensure a balanced workload, which means each expert cannot process more than $C$ tokens. This can ensure the throughput when training and deploying the MoE model with expert parallelism, \ie distributing different experts to different GPUs. However, this will also introduce an issue, the later tokens would be dropped if the previous tokens have filled the expert. In decoder-only MoE architecture, due to the auto-regressive nature, the later tokens in a sequence may be dropped more. For instance, if one expert prefers ``\textbackslash n'' token, and a sequence starts with many ``\textbackslash n''s and also has a lot of ``\textbackslash n's in the following output generated, the expert would be filled with ``\textbackslash n'' tokens quickly and all other tokens appeared later, which should be assigned to this expert, would be dropped. To verify this, we visualize the ratio of tokens dropped at different position IDs. As shown in Figure~\ref{fig:position_drop_dataset}, the general pre-training datasets, \eg RedPajama and TheStack achieved balanced token assignment, only having a small proportion of tokens dropped, even for the Position ID after 1500. However, for multi-lingual and instruction-following datasets, a large ratio of tokens is dropped. We suggest the reason is, as we discussed above, the routing decision is fixed at the early stage of training and does not change anymore, so the load balance is also achieved based on the pre-training dataset. The instruction following data can be seen as a type of out-of-domain (OOD) data of the MoE router, which would induce an unbalanced token assignment so that many tokens appearing later would be dropped.

\textbf{Can supervised fine-tuning with instruction-following data alleviate this Drop-towards-the-End issue?} Since the Drop-towards-the-End issue is mainly caused by the OOD data, it is natural to think and study whether it is possible to convert the instruction-following data to in-domain data by tuning MoE with the instruction dataset. Therefore, we compare the models before and after supervised fine-tuning in Figure~\ref{fig:position_drop}. We can see the models do not have a significant difference in the Drop-towards-the-End issue. This matches well with our insight above, \ie the routing behavior learned and fixed at the very early stage of LLM pre-training.

\subsection{Study Other MoE Models}

\textbf{Background} In this section, we investigate whether the issues we found above exist in other MoE-based LLMs, \ie Mixtral and DeepSeek-MoE. Both Mixtral and Deepseek-MoE are trained by dropless token routing, which means that these models would not drop the token even if the workload of different is unbalanced. This design is fine if our model is not that large after applying some implementation tricks like Megablock~\citep{gale2023megablocks}, which can handle the imbalanced workload better if the experts are on the same GPU. However, implementation tricks like Megablock cannot work efficiently when there is only one expert on the single GPU, and unfortunately, this happens for very large MoE LLM (e.g. one GPT-style MoE with over 2T parameters). Considering that the GPU memory size is not growing as fast as before, having a balanced workload for each expert is still extremely important for efficient large MoE model training.

\begin{figure}
\begin{minipage}[b]{0.48\linewidth}
  \begin{center}
    \includegraphics[width=1.0\textwidth]{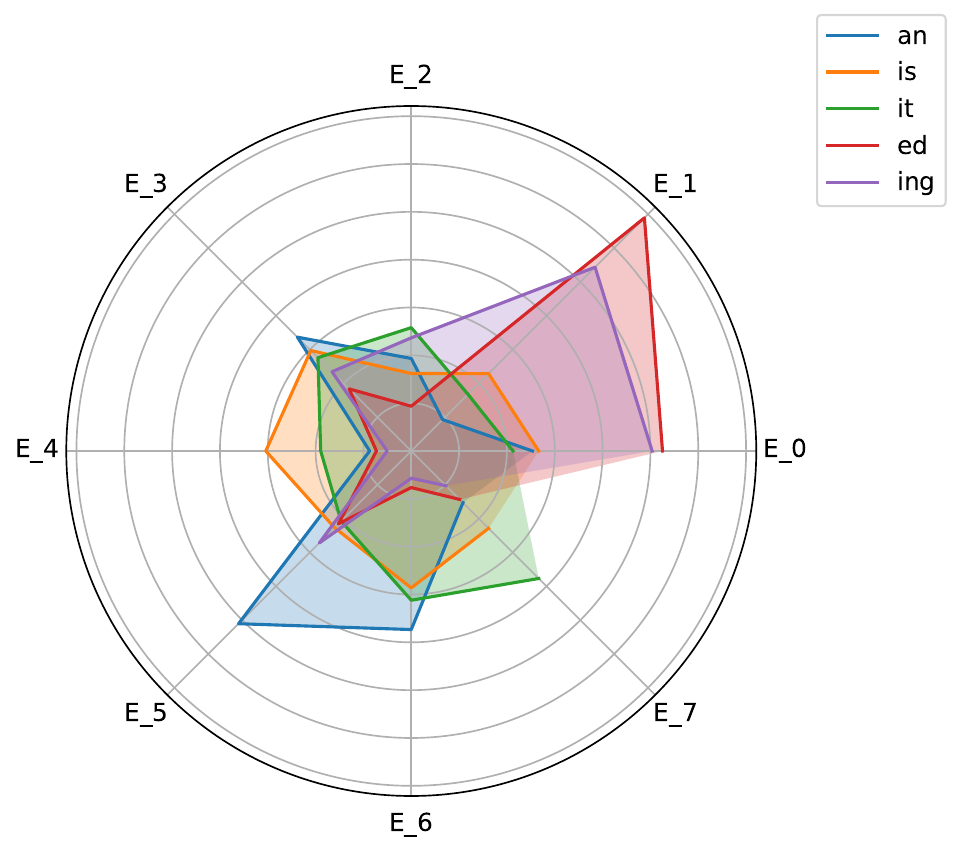}
  \end{center}
  \subcaption{Mixtral-8$\times$7B}\label{fig:token_id_specialization_mixtral}
\end{minipage}
\begin{minipage}[b]{0.48\linewidth}
  \begin{center}
    \includegraphics[width=1.0\textwidth]{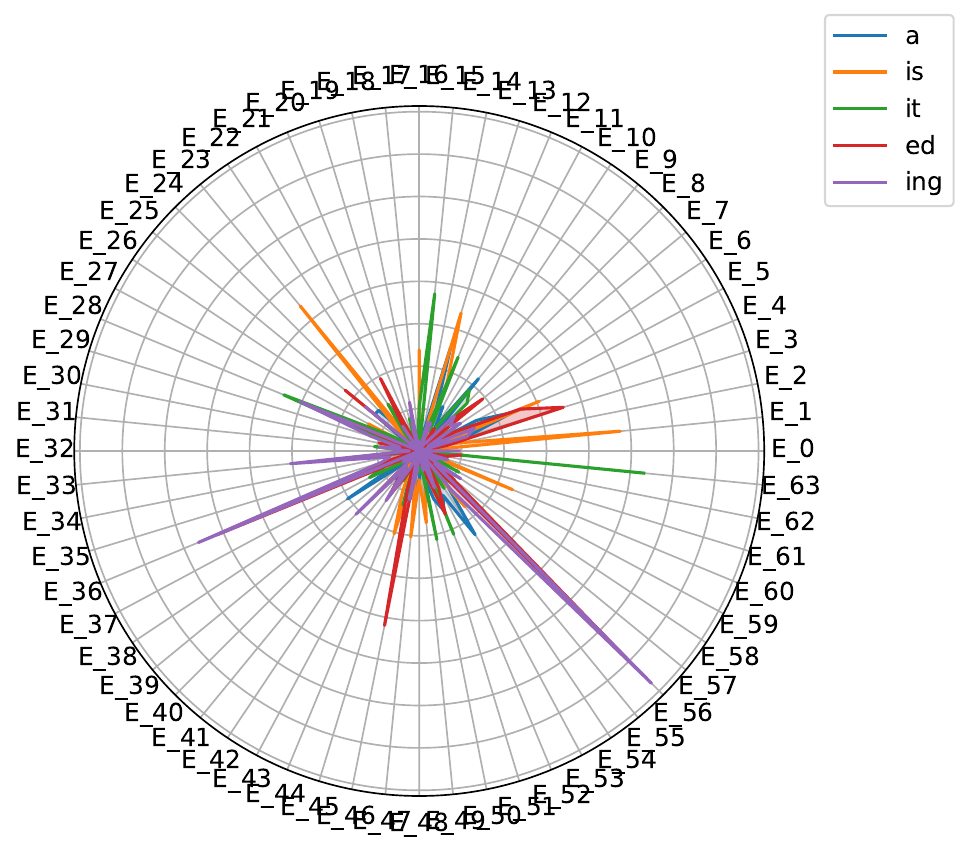}
  \end{center}
  \subcaption{Deepseek-MoE-16B}\label{fig:token_id_specialization_deepseek}
\end{minipage}
\caption{Visualization of the routing decision at different Token IDs.}
\end{figure}

\textbf{Context-Independent Specialization} We visualize the token ID specialization of Mixtral and Deepseek-MoE in Figure~\ref{fig:token_id_specialization_mixtral} and \ref{fig:token_id_specialization_deepseek}. We found, similar to OpenMoE, Deepseek-MoE has a clear Context-Independent Specialization, but Mixtral doesn’t have that. We suggest that the reason is, according to this blog\footnote{\url{https://x.com/tianle_cai/status/1734188749117153684?s=20}}, Mixtral is probably finetuned based on the Mistral-7B dense checkpoint, i.e. MoE upcycling~\cite{komatsuzaki2023sparse}, instead of training from scratch like deepseek-MoE and OpenMoE. Since the experts in Mixtral are very similar, it makes sense that there is a relatively weak specialization in their MoE model, and at the same time, since the model has learned high-level semantics when converting dense LLM to MoE LLM, it is less likely to develop Context-Independent Specialization. Therefore, we suggest that Context-Independent Specialization is an issue only for training MoE from scratch. However, as we discussed in our paper, MoE is more efficient during training than inference, it is still highly desirable to study how to avoid Context-Independent Specialization when training MoE from scratch or converting dense LLM to MoE at the early stage of training. One feasible solution can be that, first train a dense half-cooked LLM (maybe using 20\% pretraining tokens or so), and then convert the dense LLM to MoE via MoE upcycling. We can then train the MoE with 80\% of the training tokens left to ensure a better cost-effectiveness trade-off.

\textbf{Early Routing Learning} Since Mixtral and Deepseek-MoE have no open-sourced intermediate checkpoints, we cannot study this issue on these models.

\begin{figure}
\begin{minipage}[b]{0.48\linewidth}
  \begin{center}
    \includegraphics[width=1.0\textwidth]{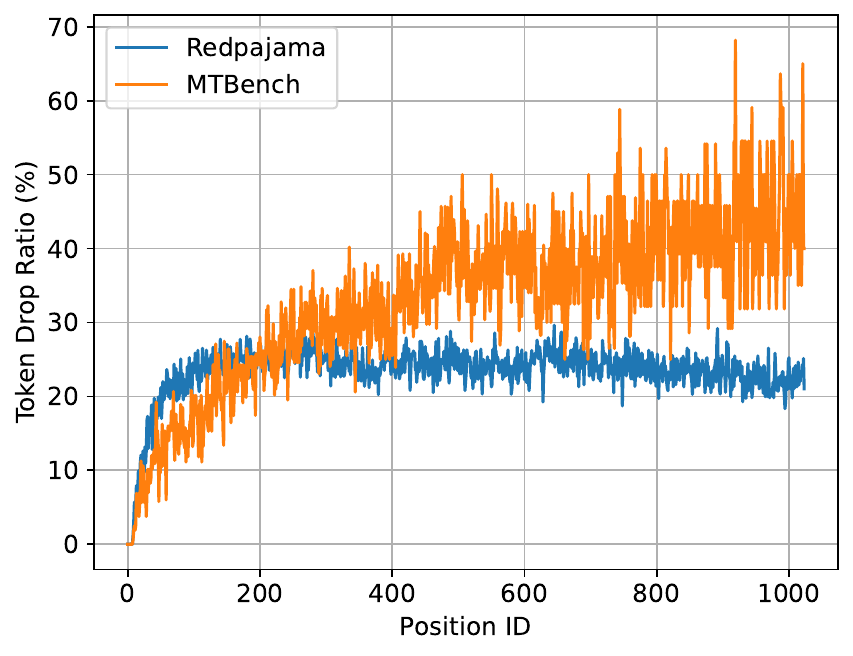}
  \end{center}
  \subcaption{Mixtral-8$\times$7B}\label{fig:position_drop_dataset_mixtral}
\end{minipage}
\begin{minipage}[b]{0.48\linewidth}
  \begin{center}
    \includegraphics[width=1.0\textwidth]{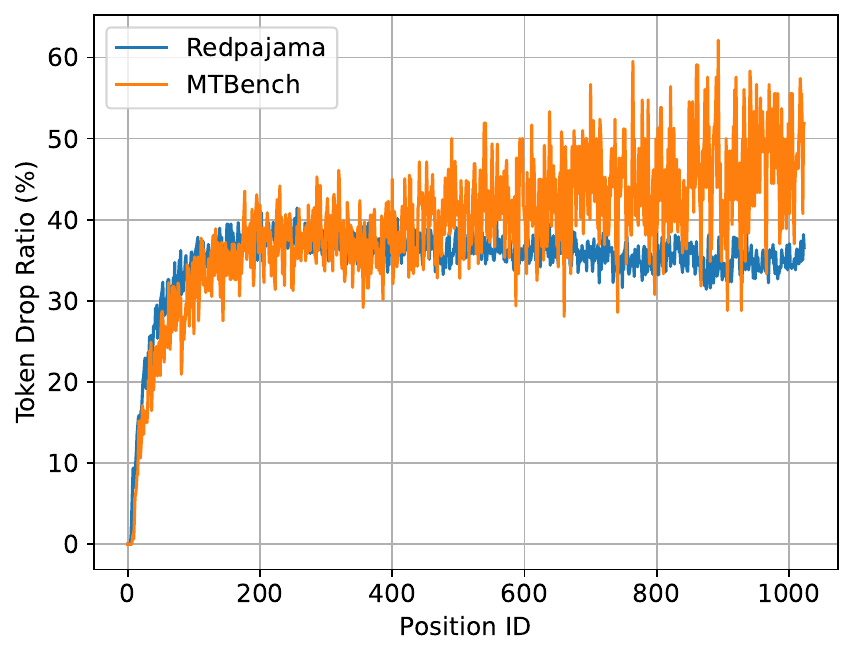}
  \end{center}
  \subcaption{DeepSeek-MoE-16B}\label{fig:position_drop_deepseek}
\end{minipage}
\caption{Comparing the ratio of tokens dropped at different position IDs.}
\end{figure}

\textbf{Drop-towards-the-End} As mentioned before, Mixtral and Deepseek-MoE have no token drop mechanism. However, this is not friendly to expert parallelism, especially for very large MoE-LLM with trillion-level parameters, although it is okay if the model is relatively small (<100B). Therefore, we still study whether there is a Drop-towards-the-End issue in Mixtral and Deepseek-MoE by manually adding a token drop mechanism when there are too many tokens routed to an expert. As shown in Figure~\ref{fig:position_drop_dataset_mixtral} and Figure~\ref{fig:position_drop_deepseek}, there is a clear token drop at the later tokens in the input sequences, which means the Drop-towards-the-End is an issue for all these MoE LLMs.

\section{Rethinking OpenMoE}\label{sec:rethink}

Working on this project is a long journey for authors. We indeed made some mistakes during design and development, but we also achieved some new insights in the analysis. We thus write down everything we found without any reservation in this paper to help future practitioners. Then, in this section, we discuss how to train a better model in the future, which are the most important takeaways of our work.

\textbf{How much code shall we use?} To be honest, we do not have a very precise answer. Conducting an ablation study is extremely expensive because of the cost of pre-training LLM at scale. The conclusion may also strongly depend on the model size and data quality. However, according to our observation, over 50\% code looks too aggressive which may harm the abilities on text tasks, but considering the importance of writing code, we suggest using around 30\% code as we used in OpenMoE-34B/32E.

\begin{table}[t]
\small
\centering
\caption{Compare umT5 tokenizer and LLaMA tokenizer on the subsets extracted from different datasets. Vocab used denotes the number of token IDs activated when tokenizing the whole subset. The umT5/LLaMA means, when tokenizing the same subset, the ratio of the number of tokens generated by umT5 and LLaMA.}
\label{tab:tokenizer-comparison}
\begin{tabular}{l|l|rr|rr|c}
\toprule
\multirow{2}{*}{\textbf{Dataset}} & \multirow{2}{*}{\textbf{Subset}} & \multicolumn{2}{c|}{\textbf{LLaMA Tokenizer}} & \multicolumn{2}{c|}{\textbf{umT5 Tokenizer}} & \multirow{2}{*}{{\textbf{umT5/LLaMA}}} \\
 & & \multicolumn{1}{c}{\textbf{\#Tokens}} & \multicolumn{1}{c}{\textbf{Vocab Used}} & \multicolumn{1}{c}{\textbf{\#Tokens}} & \multicolumn{1}{c}{\textbf{Vocab Used}} & \\
\midrule
RedPajama & arxiv & 125,339 & 8,327 & 131,059 & 8,762 & 1.046 \\
 & book & 137,972 & 11,603 & 131,072 & 15,202 & 0.950 \\
 & c4 & 28,592 & 5,439 & 26,428 & 5,554 & 0.924 \\
 & cc & 78,450 & 8,738 & 73,403 & 9,927 & 0.936 \\
 & github & 54,707 & 4,769 & 59,732 & 4,539 & 1.092 \\
 & stackexchange & 40,659 & 4,714 & 43,195 & 4,317 & 1.062 \\
 & wikipedia & 37,406 & 7,179 & 30,555 & 8,748 & 0.817 \\ \midrule
TheStack & assembly & 49,143 & 3,066 & 50,738 & 3,130 & 1.032 \\
 & blitzmax & 78,259 & 4,200 & 80,658 & 4,209 & 1.031 \\
 & java & 64,236 & 4,229 & 69,902 & 3,905 & 1.088 \\
 & python & 66,243 & 5,095 & 70,795 & 4,799 & 1.069 \\ \midrule
MTBench & writing & 6,062 & 1,700 & 5,786 & 1,535 & 0.954 \\
 & roleplay & 4,309 & 1,291 & 4,076 & 1,172 & 0.946 \\
 & reasoning & 2,369 & 478 & 2,309 & 429 & 0.975 \\
 & math & 5,163 & 290 & 5,154 & 282 & 0.998 \\
 & coding & 4,955 & 651 & 5,256 & 631 & 1.061 \\
 & extraction & 7,058 & 1,376 & 6,817 & 1,234 & 0.966 \\
 & stem & 4,783 & 1,151 & 4,527 & 1,039 & 0.946 \\
 & humanities & 6,398 & 1,451 & 5,946 & 1,320 & 0.929 \\ \midrule
Multi-lingual & ar & 256,952 & 187 & 88,406  & 8,037 & 0.344 \\
~~~~~TED & de & 103,270 & 4,880 & 80,593 & 8,470 & 0.780 \\
 & es & 101,212 & 4,745 & 78,713 & 8,519 & 0.778 \\
 & fr & 115,057 & 5,156 & 95,978 & 8,164 & 0.834 \\
 & he & 242,446 & 239 & 86,891 & 4,074 & 0.358 \\
 & it & 109,591 & 4,593 & 84,201 & 8,833 & 0.768 \\
 & ja & 144,825 & 931 & 63,491 & 6,860 & 0.438 \\
 & ko & 257,107 & 596 & 106,770 & 2,736 & 0.415 \\
 & nl & 102,703 & 4,234 & 75,084 & 7,540 & 0.731 \\
 & ru & 107,144 & 2,502 & 74,445 & 9,658 & 0.695 \\
 & zh-cn & 149,581 & 1,058 & 88,107 & 3,611 & 0.589 \\
 & zh-tw & 173,415 & 1,107 & 93,693 & 3,619 & 0.540 \\
 \bottomrule
\end{tabular}
\end{table}

\textbf{Tokenizer Selection} Our large tokenizer vocabulary introduces computation overhead at the last output layer after Transformer blocks. Although this overhead would become relatively small after scaling the Transformer model up, it is still valuable to make the tokenizer selection smarter. We conduct a quantitative analysis of the tokenizer with the datasets we used in Section~\ref{sec:analysis}. As shown in Table~\ref{tab:tokenizer-comparison}, umT5 tokenizer is indeed much better than LLaMA tokenizer on the multi-lingual dataset, especially on the low-resource language. It is also slightly better than LLaMA on the instruction-following data. However, it did not match well with our expectation that it could save more tokens for the code data. In addition, we observe that the token usage in both tokenizers is extremely long-tail distributed, which indicates that there is a large room to improve the tokenizer and following algorithms. As we know, learning from long-tailed data is hard~\cite{zhang2023deep}. Since we only have a little multi-lingual data in our pre-training data mixture, the computation cost of predicting the logits of those low-resource tokens is wasted. Based on our sub-optimal choice, we also need a solid tokenizer benchmark, which would help people evaluate tokenizers systematically. And we can then pick the best tokenizer before training the model.

\textbf{More Efficient MoE Architecture} According to our observation, MoE routing is almost context-independent (\ie Context-independent Specialization), we suggest that we can (1) remove the trainable router after warmup stage; (2) adopt parallel Transformer layer~\cite{chowdhery2022palm,gpt-j} computing FFN layer based on the input directly instead of using the output of attention layer; (3) overlapping the attention layer computation and MoE layer all-to-all communication. (1) and (3) will improve the hardware utilization and (2) can enable (3) without performance drop when scaling up~\cite{chowdhery2022palm}. 

\textbf{Mix instruction-following data during pre-training warm-up to control load balance and alleviate Drop-towards-the-End.} According to our results on multi-turn MT-Bench, it is very important to alleviate the Drop-towards-the-End issue. To this end, the key is to make the MoE achieve load balance on instruction-following data. Again, since the MoE learns and fixes the routing behavior at the early stage of pre-training, a straightforward solution is mixing the instruction-tuning data into the pre-training corpus during warm-up. This data mixing is not to align the model to learn how to follow instructions. Instead, we hope the model achieves the balanced token routing on instruction-tuning data, which paves the way to our final usage case of LLMs.

\section{Conclusion}

In this work, we explore how to train MoE for open-sourced communities. We achieved positive results that verified the effectiveness of MoE-based LLM in the post-ChatGPT stage. We disclosed all details, and our model is fully reproducible with the open-sourced code and data. More importantly, we conducted an in-depth analysis on our MoE-based LLM and found important ``Context-independent Specialization'' ``Early Routing Learning'' and ``Drop-towards-the-End''. We also rethink the mistakes we made and propose possible solutions for future developers. We sincerely hope this work can help the open-source community have a better understanding of MoE models. All the best!


\printbibliography

\newpage
\appendix
\noindent\textbf{\Large Appendix}

\section{Frequent Asked Questions}

We list the potentially frequently asked questions and the point-to-point5 answers as follows:

\subsection{Why not show the token specialization of the checkpoints at the warmup stage?} We did not expect that the routing would be learned and fixed so early. During training, due to limited storage quota, we only keep the checkpoints every 200B tokens.

\subsection{Why not compare with advanced open MoE models like Mixtral and DeepSeek-MoE?} First, our model was announced and released over 4 months earlier than Mistral and even more than DeepSeek-MoE. Second, different from models used in-house training data, our model is fully transparent. We also disclose all details and code to ensure everyone can train a comparable OpenMoE model from scratch.

\subsection{Why not use MoE upcycling?} MoE is more efficient in training instead of inference, because of better parallelism induced by large batch size. Building MoE on top of dense LLMs is a smart and faster way to get an MoE model, but not a more efficient way from a long-term view. Instead, maybe distilling MoE into a dense model~\cite{xue2022one} would be helpful if there is little performance drop.

\subsection{Why not use AdamW optimizer and Cosine Learning Rate Schedule?} We applied Adafactor optimizer and Inverse Square Root learning rate schedule following ST-MoE~\cite{zoph2022st}. We tried AdamW Optimizer but found that would introduce unstable issues (\ie NAN loss) frequently, which may introduce a significant amount of hyper-parameter sweep. Considering the limited computational resources we have, we decide to simply follow the well-studied learning rate schedule from ST-MoE~\cite{zoph2022st}.

\subsection{Why not use better and larger datasets?} When launching this project in 2023 May, there were only a few available open-source pre-training datasets. However, the scale and quality of open-sourced pre-training datasets are getting better. For instance, \citet{dolma} released 3T tokens with careful cleaning. \citet{together2023redpajama} also released a huge dataset with 30T tokens in total. We believe training on the future better data will improve the LLM performance generally by a large margin.

\newpage

\section{Hyper-parameters}\label{sec:hyper-param}

\begin{table}[h]
\centering
\tiny
\caption{Model Configurations. $H$ is the hidden size. ``Layout'' means the way of using the MoE layer. For instance, ``Every 4'' means we use one MoE layer for every 4 transformer blocks. $H_{\mathrm{FFN}}$ is the FFN intermediate size. $N_{\mathrm{Head}}$ and $H_{\mathrm{Head}}$ are the number of attention heads and attention head dimensions. $L$ is the number of layers. \#Param is the total parameters. \#ActParam is the number of parameters we used to process each token in Transformer blocks. \#ActParam w/ E is the sum of the \#ActParam and the number of parameters in the token embedding layer.}
\label{tab:model_configurations}
\begin{tabular}{l|ccccccccc}
\toprule
\textbf{Model} & \textbf{Layout} & \textbf{$H$} & \textbf{$H_{\mathrm{FFN}}$} & \textbf{$N_{\mathrm{Head}}$} & \textbf{$H_{\mathrm{Head}}$} & \textbf{$L$} & \textbf{\#Param} & \textbf{\#ActParam w/ E} & \textbf{\#ActParam} \\
\midrule
OpenMoE-Base/16E & Every 4 & 768 & 3072 & 12 & 64 & 12 & 650M & 339M & 142M \\
OpenMoE-8B/32E & Every 6 & 2048 & 8192 & 24 & 128 & 24 & 8.7B & 2.6B & 2.1B \\
OpenMoE-34B/32E & Every 4 & 3072 & 12288 & 24 & 128 & 32 & 34B & 6.8B & 6.0B \\ \midrule
TinyLLaMA & - & 2048 & 5632 & 32 & 64 & 22 & 1.0B & 1.0B & 0.9B \\
OpenLLaMA-3B & - & 3200 & 8640 & 32 & 64 & 26 & 3.0B & 3.0B & 2.9B \\
LLaMA-7B & - & 4096 & 11008 & 32 & 128 & 32 & 6.6B & 6.4B & 6.4B \\
\bottomrule
\end{tabular}
\end{table}

For OpenMoE-8B/32E, we set the head dimension as 128 instead of 64, which may be too large for a model using 2B activated Transformer parameters. We suggest that using 64 may induce a better cost-effectiveness trade-off than ours. For the number of parameters in the table above, since most parameters in Transformer blocks are from attention layer and FFN layer, we only account the trainable parameters from these two for simplicity.

\begin{table}[h]
\caption{OpenMoE training hyper-parameters.}
\label{tab:openmoe_config}
\centering
\begin{tabular}{l|ccc}
\toprule
 & \textbf{Base/16E} & \textbf{8B/32E} & \textbf{34B/32E} \\
\midrule
Optimizer & \multicolumn{3}{c}{Adafactor}  \\
Batch Size & 128 & 2048 & 2048 \\
Training Steps & 500K & 500K & 100K \\
Peak Learning Rate & \multicolumn{3}{c}{0.01}   \\
Learning Rate Schedule &  \multicolumn{3}{c}{Inverse Square Root Decay} \\
Warmup Steps & \multicolumn{3}{c}{10K}  \\
Sequence Length & \multicolumn{3}{c}{2048}  \\
Load Balance Loss Weight & \multicolumn{3}{c}{0.01}  \\
Z-Loss Weight & \multicolumn{3}{c}{0.001}  \\
Router Z-Loss Weight & \multicolumn{3}{c}{0.0001} \\
\bottomrule
\end{tabular}
\end{table}

Different from existing LLMs trained with AdamW, we used Adafactor, a more memory-efficient optimizer. Although it performs slightly worse than AdamW with the same training steps, the memory efficiency enables us to use less model parallelism and more data parallelism. In this case, using Adafactor makes our training cheaper than using AdamW to train the same model on the same data. However, we highlight that the margin of this gap is unclear because it highly depends on the hardware and model size. For our infrastructure, \ie TPUv3, this gap should be relatively larger due to the limited on-chip memory (16 GB per core).

\newpage

\section{Related Work}

\subsection{Before OpenMoE}

MoE is not new. One representative early effort is, \citet{shazeer2017outrageously} embed the MoE layer into a recurrent language model. 
Due to the scalability of Transformer architecture, GShard~\cite{lepikhin2020gshard} integrates MoE into Transformer layer and uses expert parallelism to train MoE-based Transformer at scale. 
Switch Transformer~\cite{fedus2021switch} is the earliest open-source MoE-based LM to our best knowledge, which used encoder-decoder architecture and trained with C4~\cite{2020t5} dataset. 
Due to the success of Switch Transformer on large-scale pre-training, MoE got more attention, and more advanced routing algorithms were invented. 
For instance, BASE Layers~\cite{lewis2021base} formulates token-to-expert allocation as a linear assignment problem, allowing an optimal assignment in which each expert receives an equal number of tokens. 
\citet{roller2021hash} simply modifies the feedforward layer to hash to different sets of weights depending on the current token and achieves promising results compared to learning-based routing.
Different Token-based routing above, \citet{zhou2022mixture} propose to let experts select their favorite tokens, \ie Expert-Choice Routing. Expert-choice Routing achieves more balanced token assignment and better cost-effectiveness trade-off.

Beyond the routing algorithm, there is also some work focusing on scaling MoE efficiently. \citet{artetxe2021efficient} trained their MoE models mainly on the datasets used in RoBERTa~\cite{liu2019roberta} and CC100~\cite{wenzek-etal-2020-ccnet} (112B tokens in total).
GaLM~\cite{du2022glam} further scale decoder-only MoE model with an in-house high-quality dataset with 1.6T tokens.
Brainformer~\cite{zhou2023brainformers} proposes an evolutionary search to discover MoE attributes, \eg the best way to interleave layers and layer capacities, when to fuse layers, and when to specialize layers with MoE modules and show its effectiveness at different scales.

In addition to language modeling, Vision Transformer (ViT)~\cite{dosovitskiy2020image} can also be enhanced by MoE architecture. ViT-MoE~\cite{riquelme2021scaling} verifies the scalability of MoE on ViT models. WideNet~\cite{xue2022go} shares MoE-based Transformer blocks with individual layer normalization to achieve better parameter efficiency. SoftMoE~\cite{puigcerver2023sparse} further improves the routing algorithm by applying soft token selection, which not only keeps the efficiency but also stabilizes the routing gradient. There are also some efforts devoted to include MoE into non-Transformer architecture, \eg Sparse-MLP~\cite{lou2021cross} for computer vision and s-MoE for language modeling~\cite{yu2022efficient}.

\subsection{After OpenMoE}

\begin{table}[h]
\centering
\caption{Open-sourced MoE LLMs timeline. We use the model release date as the key to sort the open-souced MoE LLMs. Dataset Size is the number of tokens in the pre-training dataset, \ie the number of tokens for one epoch. LLaMA-MoE is continued pre-trained on off-the-shelf LLaMA family models. We account its continue training dataset only.}
\label{tab:model_release_dates}
\begin{tabular}{llll}
\toprule
\textbf{Model Name} & \textbf{Dataset Size} & \textbf{Reproducible} & \textbf{Release Date} \\
\midrule
Switch Transformer~\cite{fedus2021switch} & 156B & Yes & Feb 2021 \\
Meta-MoE~\cite{artetxe2021efficient} & 112B 
 & Yes & Dec 2021 \\ \midrule
OpenMoE (Ours) & 1.1T & Yes & Aug 2023 \\ \midrule
Mixtral of Experts~\cite{jiang2024mixtral} & Unknown & No & Dec 2023 \\
LLaMA-MoE~\cite{llama-moe-2023} & 200B & Yes & Dec 2023 \\
DeepSeek-MoE~\cite{dai2024deepseekmoe} & 2T & No & Jan 2024 \\
\bottomrule
\end{tabular}
\end{table}

We released our model and implementation much earlier than writing this report. As shown in Table~\ref{tab:model_release_dates}, after our release, there are some partially open-sourced models released, \eg Mixtral~\cite{jiang2024mixtral} and Deepseek-MoE~\cite{dai2024deepseekmoe}. As we known, these models are significantly better in terms of final results. However, since these models are trained with in-house data, we have no idea about how things happened. We believe, although our results are not that amazing, the fully open-sourced nature and the in-depth analysis are both meaningful for the community.

\newpage

\section{BigBench-Lite Results}

\begin{table}[ht]
\centering
\caption{Detailed BigBench-Lite results. Note that BIG-G-sparse 8B is an MoE model with 60B parameters in total.}
\begin{tabular}{lS[table-format=2.2]S[table-format=2.2]S[table-format=2.2]S[table-format=2.2]}
\toprule
Model & \multicolumn{1}{c}{BIG-G 8B} & \multicolumn{1}{c}{BIG-G-sparse 8B} & \multicolumn{1}{c}{GPT-3 6B} & \multicolumn{1}{c}{\textbf{OpenMoE-8B}} \\
\midrule
auto\_debugging & 0.0 & 0.0 & 0.0 & 17.65 \\
bbq\_lite\_json & 58.63 & 46.13 & 49.85 & 42.67 \\
code\_line\_description & 4.66 & 2.44 & 20.18 & 2.44 \\
conceptual\_combinations & -2.16 & 1.07 & -3.36 & 0.81 \\
conlang\_translation & 31.38 & 33.25 & 37.92 & 36.93 \\
emoji\_movie & 3.75 & 7.5 & -5.0 & 3.75 \\
formal\_fallacies\_syllogisms\_negation & 0.78 & -0.39 & -0.8 & -0.56 \\
hindu\_knowledge & 12.44 & 8.63 & 19.29 & 16.24 \\
known\_unknowns & -34.78 & -4.35 & -8.7 & -13.04 \\
language\_identification & 1.39 & -0.33 & 1.66 & 1.77 \\
linguistics\_puzzles & 0.0 & 0.0 & 0.0 & 0.05 \\
logic\_grid\_puzzle & -2.45 & 0.01 & -0.28 & 0.89 \\
logical\_deduction & 1.38 & 4.2 & 1.05 & 0.09 \\
misconceptions\_russian & -34.69 & -38.78 & -34.69 & -38.78 \\
novel\_concepts & 10.16 & 14.06 & 17.97 & 6.25 \\
operators & 10.48 & 16.67 & 20.0 & 20.48 \\
parsinlu\_reading\_comprehension & 0.0 & 0.0 & 0.0 & 11.97 \\
play\_dialog\_same\_or\_different & 12.5 & 4.69 & -3.8 & 1.1 \\
repeat\_copy\_logic & 0.0 & 6.25 & 0.0 & 3.12 \\
strange\_stories & -7.15 & -4.77 & 9.54 & 14.52 \\
strategyqa & 7.23 & 8.4 & -3.8 & 3.36 \\
symbol\_interpretation & 6.06 & 0.13 & 4.17 & 2.65 \\
vitaminc\_fact\_verification & 6.25 & 1.27 & -3.2 & 21.34 \\
winowhy & 4.69 & 5.27 & 11.6 & 10.14 \\ \midrule
Average & 3.77 & 4.63 & 5.40 & \textbf{6.93} \\
\bottomrule
\end{tabular}

\label{tab:bblite-results}
\end{table}

\newpage

\section{Routing Decision Standard Deviation}\label{sec:routing_std}

    

\begin{figure}[h]
\centering
\centering
\includegraphics[width=0.5\textwidth]{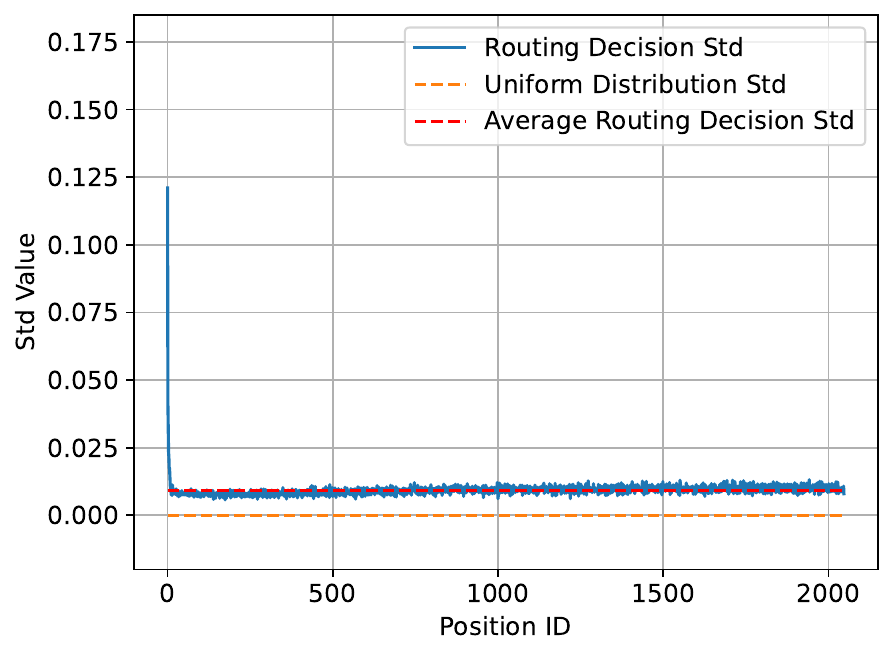}
\caption{The routing decision standard deviation at different position IDs.}\label{fig:position_std}
\end{figure}

\begin{figure}[h]
\centering
\includegraphics[width=0.5\textwidth]{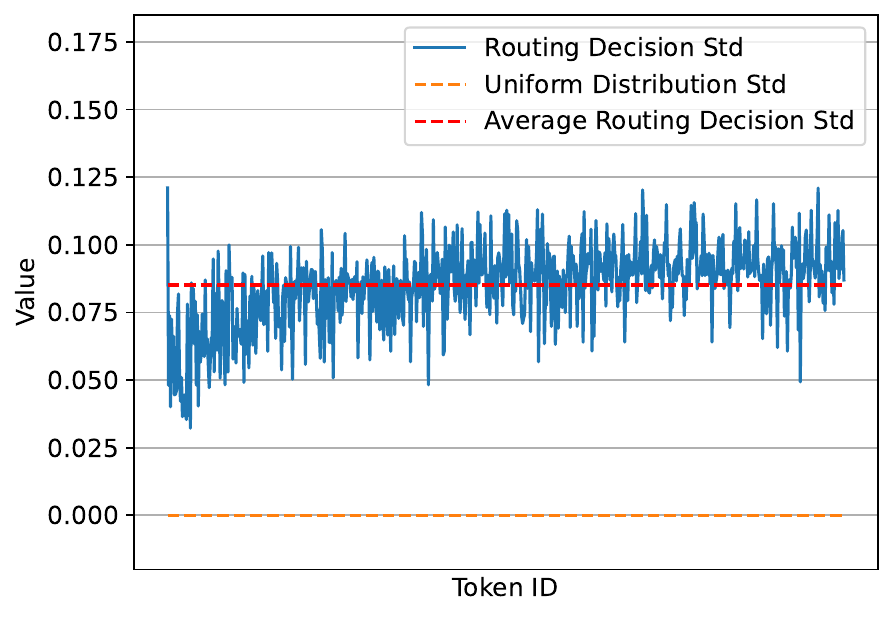}
\caption{The routing decision standard deviation at different token IDs. We only take the token IDs with over 128 tokens, because the extremely low-resourced tokens always have large routing decision standard deviation. The token IDs never appeared also have variance at all.}\label{fig:token_std}
\end{figure}

In Figure~\ref{fig:position_std} and~\ref{fig:token_std}, we can clearly see that the token IDs have a larger standard deviation on routing decisions than position IDs. Also, most token IDs have a relatively large standard deviation, which means most of the token IDs have Context-independent Routing.

\newpage

\section{Top Token Selection by Experts}\label{sec:top_token_section}

\begin{table}[h]
\centering
\caption{Top Tokens selected by each expert.}
\label{tab:top_token_table_full}
\begin{tabular}{c|l}
\toprule
Expert ID & Top Tokens \\
\midrule
0 & ``\textbackslash n', `` ` '', `` ’ '', ``s'', ``-'', ``\$'', ``y'', ``\_'', `` '', ``2'' \\
1 & ``\textbackslash n', ``1'', `` '', ``2'',  ``'\textbackslash\textbackslash'', ``S'', ``.'', ``-'', ``C'', ``\{'' \\
2 & ``in'', ``.'', ``2'', ``1'', ``0'', ``\textbackslash n', `` '', ``3'', ``\_'', ``4'' \\
3 & ``s'', ``)'', ``a'', ``\textbackslash n', ``which'', ``es'', ``);'', ``\}'', ``\textbackslash\textbackslash'', ``e'' \\
4 & ``\textbackslash n', ``.'', ``0'', ``the'', ``,``, ``\_'', ``that'', ``1'', ``as'', ``\^'' \\
5 & `` '', ``\textbackslash n', ``s'', ``2'', ``a'', ``on'', ``ter'', ``*'', ``\textbackslash\textbackslash'', ``all'' \\
6 & ``the'', ``,'', ``.'', ``a'', ``to'', ``of'', `` '', ``s'', ``de'', ``\textbackslash n' \\
7 & ``,``, ``and'', ``\textbackslash n', ``:'', ``\_'', `` '', ``0'', ``on'', ``at'', ``\{'' \\
8 & ``('', ``.'', ``that'', ``s'', `` '', ``,``, ``C'', ``which'', ``of'', ``G'' \\
9 & ``('', ``this'', ``2'', ``\textbackslash n', ``\textbackslash\textbackslash'', `` '', ``3'', ``also'', ``I'', ``1'', ``,`` \\
10 & ``\textbackslash n', ``.'', ``and'', ``\textbackslash r'', ``).'', ``;'', ``\textbackslash t'', ``:'', ``?'', ``The'' \\
11 & ``to'', ``1'', ``the'', ``2'', ``0'', ``s'', ``for'', ``t'', ``3'', ``\textbackslash n' \\
12 & ``the'', ``,``, ``\$', ``to'', ``in'', ``?'', ``as'', ``that'', ``In'', ``who'' \\
13 & ``in'', ``/'', ``0'', ``\textbackslash n', ``with'', ``-'', `` '', ``\{'', ``of'', ``2'' \\
14 & ``is'', ``.'', ``are'', ``be'', ``was'', ``s'', ``\textbackslash n', ``,``, ``has'', ``not'' \\
15 & ``of'', ``\textbackslash n', ``\_'', ``s'', `` '', ``.'', ``S'', ``the'', ``for'', ``\textbackslash\textbackslash'' \\
16 & ``cite'', ``,``, ``\textbackslash n', ``.'', ``\{'', ``s'', "'", ``ing'', ``data'', ``\textbackslash\textbackslash\textdollar'', ``\textbackslash t'' \\
17 & ``the'', ``.'', ``\textbackslash n', ``The'', ``0'', ``1'', ``as'', ``of'', ``5'', ``2'' \\
18 & ``-'', ``\{'', ``for'', ``('', ``\_'', `` '', ``\textdollar'', ``('', ``\textbackslash n', ``\}'' \\
19 & `` '', ``and'', ``in'', ``to'', ``,``, ``of'', ``or'', ``\textbackslash n', ``by'', ``\textdollar'' \\
20 & ``\textbackslash n', ``the'', ``\textdollar'', ``a'', ``0'', ``\}'', ``this'', ``1'', ``s'', ``9'', `` '' \\
21 & ``,``, ``and'', `` '', ``.'', ``\textbackslash n', ``='', ``\textbackslash t'', ``the'', `` '', ``n'' \\
22 & ``the'', ``\textbackslash n', ``)'', ``,``, ``his'', ``their'', ``s'', '"', ``,``, ``i'' \\
23 & ``.'', ``\textbackslash n', ``,``, ``*'', ``<pad>'', ``Cha'', ``i'', ``!'', ``our'', ``/'' \\
24 & ``a'', ``with'', '\}'', ``in'', ``)''', ``:'', ``an'', ``1'', ``\textbackslash n', ``at'' \\
25 & ``\textbackslash\textbackslash'', ``the'', ``.'', ``of'', ``er'', ``, '', ``s'', ``ter'', ``book'', ``model'' \\
26 & ``\textbackslash n', ``, '', ``.'', ``a'', ``<pad>'', ``s'', ``de'', ``al'', ``-'' \\
27 & ``the'', "'", ``I'', ``The'', ``, '', ``it'', ``we'', ``he'', ``a'', ``x'' \\
28 & ``, '', ``ly'', ``\{'', ``\_\{'', ``new'', ``-'', ``ed'', ``more'', ``\textbackslash n', ``d'' \\
29 & ``, '', ``.'', ``of'', ``;'', ``by'', ``,:'', ``\textbackslash n', ``to'', ``from'', ``('', \\
30 & ``\}'', ``ed'', ``d'', ``have'', ``ing'', ``, '', ``has'', ``s'', '"', ``had'' \\
31 & ``to'', ``can'', ``s'', ``of'', ``ing'', ``will'', ``not'', ``e'', ``ed'', ``would'' \\
\bottomrule
\end{tabular}
\end{table}

\end{document}